%% file: acl_latex.tex
\documentclass[11pt]{article}

\usepackage[final]{acl}

\usepackage{times}
\usepackage{latexsym}

\usepackage{pifont} 
\newcommand{\cmark}{\ding{51}}
\newcommand{\xmark}{\ding{55}}
\usepackage{svg}
\usepackage{tabularx}
\usepackage{booktabs, array, multirow, threeparttable}
\usepackage{amssymb}
\usepackage{pifont}
\usepackage{amsmath}        
\usepackage[most]{tcolorbox}
\usepackage{float}
\usepackage{algorithm}
\usepackage{algorithmic}
\usepackage{siunitx}

\usepackage[T1]{fontenc}

\usepackage[utf8]{inputenc}

\usepackage{microtype}

\usepackage{inconsolata}

\usepackage{graphicx}

\title{ToolScope: Enhancing LLM Agent Tool Use through Tool Merging and Context-Aware Filtering}

\author{
Marianne Menglin Liu \quad Daniel Garcia \quad Fjona Parllaku \\ 
\textbf{Vikas Upadhyay} \quad \textbf{Syed Fahad Allam Shah} \quad \textbf{Dan Roth}\\
Oracle AI\\
\small
\texttt{marianne.liu@oracle.com}\\
}

\begin{document}
\maketitle
\begin{abstract}
Large language model (LLM) agents rely on external tools to solve complex tasks. However, real-world toolsets often contain semantically redundant tools with overlapping names and descriptions, introducing ambiguity and degrading tool selection performance. In addition, LLMs face strict input context limits, which prevent the agent from efficiently considering a large number of tools per query. To address these challenges, we propose ToolScope, a novel approach which contains: (1) ToolScopeMerger with Auto-Correction: automatically audits and fixes tool merges, reducing semantic redundancy in large toolsets. (2) ToolScopeRetriever, which ranks and selects only the top-k relevant tools for a given query, effectively compressing the toolset to fit within the LLM’s input window without sacrificing selection accuracy. This selective filtering directly mitigates context length constraints by ensuring that only the most relevant tools are passed to the model. We evaluate ToolScope using 3 state-of-the-art LLMs across 3 open-source tool-use benchmarks covering both single-tool and multi-tool scenarios in diverse real-world domains. Experimental results show a substantial increase of 8.38\% to 38.6\% in tool selection accuracy, demonstrating ToolScope’s effectiveness in enhancing LLM tool-use capabilities.
\end{abstract}

\section{Introduction}

Tool learning \cite{qu2025tool}, which refers to LLMs becoming proficient in using external tools to solve complex problems \cite{qin2023toolllm}, has emerged as a key research topic. It is divided into four stages: task planning, tool selection, tool calling, and response generation \cite{qu2025tool}. Among these, tool selection remains a critical yet underdeveloped sub-task \cite{yang2023gpt4tools, qin2023toolllm}, as LLMs often fail to choose the most appropriate tools for a given query. Two major factors that contribute to this are: (1) overlapping tool descriptions, which introduce ambiguity that reduces both retrieval and selection accuracy \cite{openai2024agents, lumer2024toolshed, huang2023metatool} and (2) the limited input context of LLMs, which constrain the model's ability to reason effectively over large toolsets \cite{huang2023metatool}. As toolsets continue to grow in size and complexity, these issues become more pronounced. Therefore, it is essential to address both tool overlap and context limitations for improved tool selection performance.

\begin{figure}[t]
  \centering
  \includegraphics[width=\columnwidth]{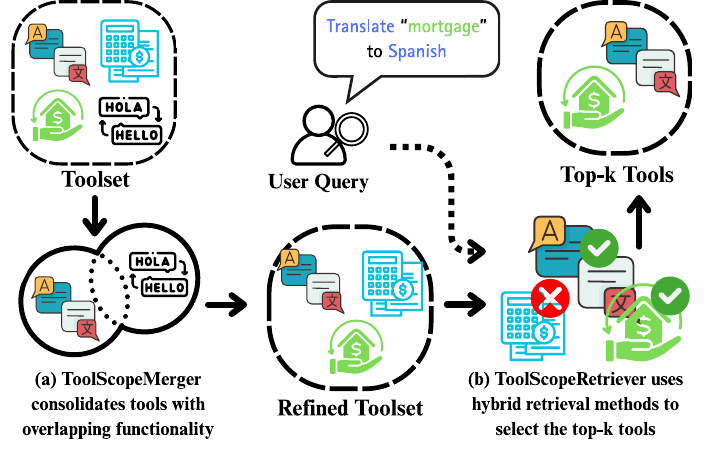}
  \caption{A simplified overview of using ToolScope for a tool selection task.}
  \label{fig:overview4}
\end{figure}


Previous research has explored several directions for improving tool selection. Tuning-free methods \cite{qu2025tool} have improved tool documentation with frameworks such as DRAFT \cite{qu2024exploration} and EASYTOOL \cite{yuan2024easytool}. Retrieval-based methods have investigated “term-based” approaches such as BM25 \cite{robertson2009probabilistic}, “semantic-based” methods such as CRAFT \cite{yuan2023craft}, and off-the-shelf embeddings \cite{qu2025tool} as the retrieval method to improve tool selection. 

However, these approaches face several key limitations: (1) They only improve individual tool documentations, failing to address semantic overlap across the toolset. Methods that do address this overlap rely on manual tool merging. (2) Limited studies have evaluated the impact of hybrid retrieval methods on tool selection. (3) No prior work has combined automated tool merging with hybrid retrieval to simultaneously address tool overlap and context limitations.

We propose ToolScope, a two-part solution that automatically merges tools through ToolScopeMerger and retrieves the most relevant tools for a query using ToolScopeRetriever. This addresses the two main challenges presented: overlapping tool descriptions and context length limitations. 

ToolScopeMerger automatically cleans raw toolsets in three stages. First, the system identifies candidate overlaps through semantic similarity. Then, it validates relationships with an LLM to ensure precise merges. Finally, it consolidates clusters with automated correction to produce a merged toolset that preserves tool capabilities while improving downstream retrieval by offering a smaller selection of tools.

ToolScopeRetriever addresses single and multi-tool queries through a hybrid retrieval strategy. This is followed by reranking system to ensure a fair, contextually relevant selection. By combining semantic and exact matches, it improves results for retrieval and tool selection accuracy.

In summary, our contributions are: (1) \textbf{ToolScopeMerger}, a graph-based, automated and scalable framework with Auto-Correction that merges semantically similar tools to reduce tool overlap in large toolsets. (2) \textbf{ToolScopeRetriever}, a hybrid retrieval system that combines sparse and dense scores to reduce context length. (3) A joint framework combining (1) and (2) significantly improves retrieval and selection accuracy across open-source benchmarks and models, boosting tool selection accuracy by 34.6\% on Seal-Tools, 38.6\% on UltraTool, and 8.8\% on BFCL (Table~\ref{tab:tool_selection_models_first}).

\begin{figure*}
  \centering
  \includegraphics[width=1\textwidth]{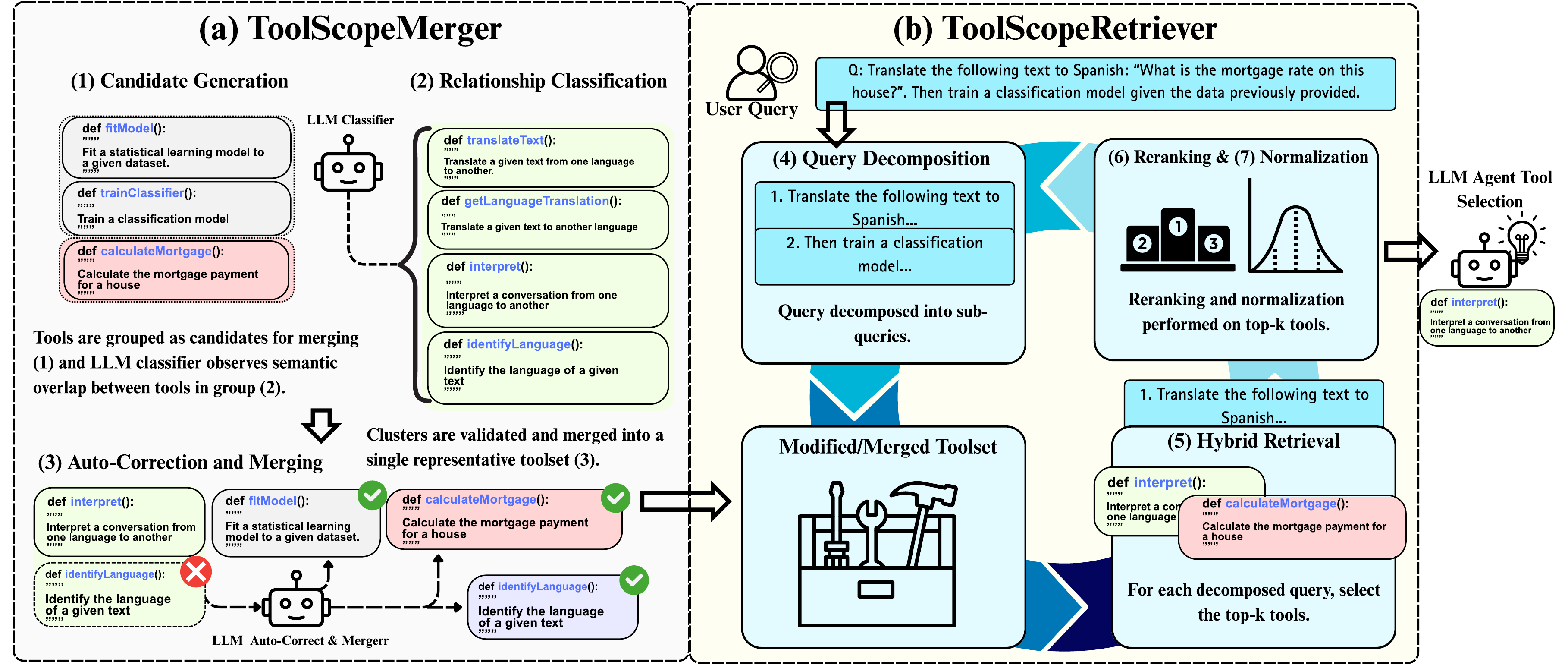}
  \caption{An overview of ToolScope which consists: (a) ToolScopeMerger, steps including: (1) Candidate Generation, (2) Relationship Classification, (3) Auto-Correction and Merging), (b) ToolScopeRetriever, steps including: (4) Query Decomposition, (5) Retrieval, (6) Reranking and (7) Normalization. Retrieved tools are then presented to LLM Agent for tool selection.}
  \label{fig:toolscope_overview}
\end{figure*}

\section{Related Work}
\subsection{Tool Learning}
Tool learning \cite{qu2025tool}, which refers to LLM as being proficient in using tools to solve complex problems as humans \cite{qin2023toolllm}, can be broken down into four stages of learning: task planning, tool selection, tool calling, and response generation \cite{qu2025tool}. 

Advancements in tool learning can be divided into two categories: tuning-based and tuning-free methods \cite{qu2025tool}. Tuning-based approaches encompass supervised fine-tuning \cite{yang2023gpt4tools, liu2024toolace, shen2024llm, acikgoz2025can}, contrastive learning \cite{wu2024avatar}, reinforcement learning \cite{wang2025otc, qian2025toolrl}, and extending the modeling head with trainable tool token embeddings \cite{alazraki_2025}. However, these approaches are computationally expensive, requiring substantial model fine-tuning. Tuning-free approaches include chain-of-thought prompting \cite{inaba2023multitool, liu2024toolace}, data augmentation \cite{liu2024toolace}, and response-reasoning strategies \cite{liu2024toolace}. Tuning-free approaches present advantages that can be explored, such as working well on both open- and closed-source models \cite{qu2025tool}.
While recent work has improved tool use by enhancing individual tool descriptions \cite{huang2023metatool, qu2024exploration}, it overlooks cross-tool semantic relationships, limiting its ability to resolve redundancy and ambiguity in large toolsets. 

\subsection{Tool Overlap}
Several prior works have acknowledged the issue of tool overlap, which is defined as a query that can be solved by multiple tools \cite{huang2023metatool}, but fail to provide a concrete, autonomous solution. OpenAI’s guide to LLM Agents highlights tool overlap as one of the possible causes of LLM agents consistently select incorrect tools and suggests user keep tools distinct \cite{openai2024agents}. ToolShed \cite{lumer2024toolshed} identifies this issue of tool overlap in several notable tool selection benchmarks such as ToolBench \cite{qin2023toolllm} and ToolAlpaca \cite{tang2023toolalpaca}. Prior work mitigates tool overlap via clustering \cite{huang2023metatool} and manual curation \cite{huang2023metatool}; \cite{huang2024planningcreationusagebenchmarking}, yet no automated, scalable solution has been proposed.

\subsection{Retrieval-Based Tool Selection}
As tool libraries expand, selecting the appropriate tool increasingly resembles a retrieval task. Prior work has examined lexical methods like BM25 \cite{robertson2009probabilistic} and semantic methods such as CRAFT \cite{yuan2023craft}. Recent retrieval approaches, including RAG-Tool Fusion \cite{lumer2024toolshed}, ScaleMCP \cite{lumer2025scalemcp} incorporate Retrieval Augmented Generation (RAG) strategies through pre-retrieval, intra-retrieval, and post-retrieval phases. Those approaches leverage query reformulation, re-ranking, and retrieval-based planning to improve retrieval accuracy. However, limited studies have evaluated hybrid retrieval \cite{lumer2025scalemcp}. Finally, no study has combined automatic tool merging and hybrid tool retrieval to improve LLM agent tool selection.

\section{Methods}
\subsection{Overview of ToolScope}
ToolScope is a framework designed to optimize the performance of LLM agents by addressing the two key challenges identified in tool selection: tool overlap and context length limitation. Figure~\ref{fig:toolscope_overview} presents an overview of ToolScope.

\subsection{ToolScopeMerger}
\label{sec:merger}
ToolScopeMerger is a graph-based framework that consolidates overlapping tools in large agent toolsets. An optional Auto-Correction stage uses an LLM validator to audit clusters, confirm equivalence, and split or remove members when errors in merging are detected.

We organize ToolScopeMerger into three stages: S1: Candidate Generation (P1: Tool Indexing), S2: Relationship Classification and Graph Formation (P2: Relationship Classification, P3: Graph Construction), and S3: Consolidation and Auto-Correction (P4: Tool Pruning, P5: Auto-Correction, P6: Toolset and Dataset Update).

\textbf{Tool Indexing.} Let the toolset be defined as \( T = \{t_1, t_2, \dots, t_n\} \). For each tool \( t_i \), its signature and  natural language  description \( d_i \) are encoded using an embedding model \( f \) to produce a dense vector representation:
\(v_i = f(d_i), \forall i \in \{1, \dots, n\}.
\)
Let \( V = [v_1, \dots, v_n] \in \mathbb{R}^{n \times d} \) denote the matrix of all embedded tools, where \( d \) is the embedding dimension. For each \( t_i \), we retrieve its top-\( k \) most similar tools based on cosine similarity, forming the candidate set:
\(
T_i^{(k)} = \{t_{i_1}, \dots, t_{i_k}\}, \ \mathrm{where} \ t_{i_j} \in T \setminus \{t_i\}.
\)
This set \( T_i^{(k)} \) contains the candidates for potential merging with \( t_i \).

\textbf{Tool Relationship Classification.} We define an LLM-based binary classifier \( M_C : T \times T \to \{0, 1\} \) to detect semantic overlap between tools. Given a tool pair, the model determines whether the two serve sufficiently similar functions to justify merging. The input format and prompting strategy for this classification are described in Appendix \ref{app:H1}. For each pair \( (t_i, t_j) \) where \( t_j \in T_i^{(k)} \), the classifier outputs a binary label indicating whether the tools are semantically equivalent, as defined in Equation~\ref{eq:match_matrix}.

\begin{equation}
M_C(t_a, t_b) = 
\begin{cases}
1 & \mathrm{if} \ t_a \sim t_b \\
0 & \mathrm{otherwise}
\end{cases}
\label{eq:match_matrix}
\end{equation}
where $\sim$ indicates semantic equivalence.

\textbf{Graph Construction.} Inspired from the Tool Graph which utilizes graph to represent relationships and dependencies between tools \cite{shen2024taskbench, liu2024toolnet}, we define an undirected pruning graph $G=(T,E)$, where each node represents a tool $t_i \in T$, and an edge $(t_a, t_b) \in E$ exists if $M_C (t_a, t_b) = 1$. We then extract the set of connected components in $G$, $C = \{C_1, C_2,..., C_a \}$, where each component $C_b \subseteq T$ contains tools that are considered semantically equivalent and therefore overlapping.

\textbf{Tool Pruning.} For each connected component \( C_b \), one representative tool \( t^{*}_b \in C_b \) is selected to serve as the canonical tool for the component. Currently, our implementation selects the tool that minimizes function name string length. We define the pruned toolset as:
\(
T' = \{t_1^*, t_2^*, \dots, t_k^*\}, \ \mathrm{where} \ k \leq n.
\)
We also maintain a mapping:
\(
\phi : T \to T', \mathrm{where} \ \phi(t) = t_j^* \mathrm{, } \ \forall t \in C_b.
\)

\textbf{Auto-Correction.}
To ensure high-precision merges, we validate and correct tool pruning automatically using an LLM validator $V$ that can be easily turned on and off. For each proposed cluster merge, $V(C_b)$ determines whether all functions satisfy semantic equivalence criteria (see Appendix \ref{app:H2}). If the cluster is valid, the merge proceeds. If the cluster is invalid, $V(C_b)$ proposes a refined splitting of the original cluster $C_b$ into more granular sub-clusters and removes members that are not semantically equivalent. 

\begin{equation}
V(C_b) = \{C'_1, \dots, C'_m\}, \quad
\mathrm{where \ each} \ C'_j \subseteq C_b
\label{eq:validator_output}
\end{equation}

This iterative correction removes the need for manual review. Final integrity checks ensure no tool appears in multiple prune dictionaries and none is simultaneously kept and pruned. This guarantees a consistent one-to-one mapping \(\phi: T \to T'\) for merging and dataset relabeling.

\textbf{Toolset and Dataset Update.} For each cluster $C_b$, an LLM $M_D$ is prompted to synthesize a new tool signature and description $d^*$ such that $M_D(C_b) = d^*$. The new signature and tool description $d^*$ unifies the functionality of the tools in $C_b$. See Appendix \ref{app:H3} for prompt details. This forms a new merged tool based around the representative tool $t^{*'}_{j} = d^*$. Given our mapping $\phi$, we update our original benchmark dataset $\beta$ by relabeling our gold responses, as shown by the Equation~\ref{eq:relabel}:


\begin{equation}
\forall (q, l) \in \beta, \; t \Longrightarrow \phi(t)
\label{eq:relabel}
\end{equation}
where $(q, l)$ is a query-response pair, and $l$ can be multiple tools. This final step ensures the new Toolset $T'$ is still compatible with the evaluation benchmark and leads to fair and accurate testing.

\subsection{ToolScopeRetriever}
Effective tool selection is essential for enhancing the capabilities of large language models (LLMs) when solving complex tasks involving external tools. We adopt a hybrid approach to tool selection that integrates both local and global reranking strategies, supporting both single-tool and multi-tool use cases.

\textbf{Single-tool selection.} Given a query $q$ and a set of candidate tools $T$, we compute a hybrid retrieval score for each tool $t \in T$ by combining sparse and dense similarity scores through a weighted average, denoted as:
\begin{equation}
s(q, t) = \alpha \cdot s_{\mathrm{dense}}(q, t) + (1 - \alpha) \cdot s_{\mathrm{sparse}}(q, t)
\label{eq:hybrid_score}
\end{equation}
The top $M$ candidates based on $s(q, t)$ are then reranked using a cross-encoder that scores each pair $(q, t)$. The tool with the highest reranker score is selected as the final output.

\textbf{Multi-tool selection.} In the multi-tool setting, each query $q$ is associated with multiple subqueries. For each subquery, we apply the same hybrid retrieval and reranking procedure. We then select the top-1 tool $t^{(1)}$ with the highest reranker score from each subquery. The remaining tools $t^{(j)}$ for $j = 2, \dots, M$ are normalized using min-max normalization:
\begin{equation}
s_{\mathrm{norm}}(t^{(j)}) = \frac{s(t^{(j)}) - s_{\min}}{s_{\max} - s_{\min} + \varepsilon}
\label{eq:score_normalization}
\end{equation}
This normalization rescales scores across subqueries, since raw reranker scores may not be directly comparable due to varying query-to-tool similarity distributions. It enables a fair global ranking of tools from different subqueries. The top-$k$ toolset is assembled by first including all top-1 selections, then iteratively adding tools with the highest normalized scores until $k$ tools are selected.

\section{Experiments}
\subsection{Experiment Setup}

\begin{table}[t]
\centering
\small
\begin{tabularx}{\linewidth}{
  >{\raggedright\arraybackslash}X   
  >{\centering\arraybackslash}p{1.8cm}  
  >{\centering\arraybackslash}p{1.5cm}  
  >{\centering\arraybackslash}p{2.0cm}  
}
\toprule
\textbf{Dataset} & \textbf{\# Queries} & \textbf{\# Tools} & \textbf{Multi-Tool?} \\
\midrule
\textbf{BFCL}       & 400  & 400  & No \\
\textbf{Seal-Tools} & 654 & 4076 & \textbf{Yes}\\
\textbf{UltraTool}  & 4814 & 1885 & \textbf{Yes} \\
\bottomrule
\end{tabularx}
\caption{Benchmark datasets used in our evaluation, including scale and presence of multi-tool scenarios.}
\label{tab:datasets}
\end{table}

\begin{table*}[t]
\centering
\small
\begin{threeparttable}
\begin{tabularx}{\textwidth}{
    l l 
    *{3}{>{\centering\arraybackslash}X}  
    *{3}{>{\centering\arraybackslash}X}  
    >{\centering\arraybackslash}X        
}
\toprule
\textbf{LLM} & \textbf{Method} &
\multicolumn{3}{c}{\textbf{Seal-Tools}} &
\multicolumn{3}{c}{\textbf{BFCL}} &
\textbf{UltraTool} \\
\cmidrule(lr){3-5} \cmidrule(lr){6-8} \cmidrule(lr){9-9}
& & CSR@5 & CSR@10 & CSR@15 & CSR@5 & CSR@10 & CSR@15 & CSR@30 \\
\midrule

\multirow{4}{*}{\textit{GPT-4o}}
  & BM25        & 0.544 & 0.593 & 0.646 & 0.850 & 0.875 & 0.870 & 0.366 \\
  & Dense       & 0.548 & 0.603 & 0.631 & 0.888 & 0.900 & 0.915 & 0.673 \\
  & ToolScope            & 0.875 & 0.896 & 0.905 & 0.912 & 0.915 & 0.915 & 0.723 \\
  & ToolScope+AutoCorrect & \textbf{0.890} & \textbf{0.921} & \textbf{0.924} & \textbf{0.938} & \textbf{0.935} & \textbf{0.940} & \textbf{0.752} \\
\midrule

\multirow{4}{*}{\textit{LLaMA-3.3-70B}}
  & BM25        & 0.527 & 0.574 & 0.618 & 0.832 & 0.848 & 0.865 & 0.364 \\
  & Dense       & 0.531 & 0.584 & 0.611 & 0.840 & 0.845 & 0.862 & 0.664 \\
  & ToolScope            & 0.883 & 0.900 & 0.910 & 0.888 & 0.892 & 0.882 & 0.729 \\
  & ToolScope+AutoCorrect & \textbf{0.889} & \textbf{0.916} & \textbf{0.922} & \textbf{0.935} & \textbf{0.942} & \textbf{0.942} & \textbf{0.759} \\
\midrule

\multirow{4}{*}{\textit{Command-R-08-2024}}
  & BM25        & 0.463 & 0.512 & 0.551 & 0.785 & 0.805 & 0.807 & 0.368 \\
  & Dense       & 0.523 & 0.569 & 0.603 & 0.802 & 0.825 & 0.822 & 0.669 \\
  & ToolScope            & 0.873 & 0.888 & 0.889 & 0.902 & \textbf{0.912} & \textbf{0.920} & 0.702 \\
  & ToolScope+AutoCorrect & \textbf{0.878} & \textbf{0.900} & \textbf{0.902} & \textbf{0.908} & 0.900 & 0.900 & \textbf{0.718} \\
\bottomrule
\end{tabularx}

\end{threeparttable}
\caption{Correct Selection Rate (CSR@$k$) for each method and LLM. 
BFCL is evaluated at $k\in\{5,10,15\}$, Seal-Tools at $k\in\{5,10,15\}$, and UltraTool at $k=30$. 
$k$ refers to the top $k$ tools retrieved and passed to the LLM agent for tool selection. Full results are shown in Table \ref{tab:csr_grouped_by_dataset} in Appendix \ref{app:C}.}
\label{tab:tool_selection_models_first}
\end{table*}

\textbf{Datasets.} Prior work has introduced datasets such as ToolACE \cite{liu2024toolace}, ToolE \cite{huang2023metatool}, Berkeley Function Calling Leaderboard (BFCL) \cite{patil2024gorilla}, NexusRaven \cite{ravenv2}, ToolBench \cite{qin2023toolllm}, RestBench \cite{song2023restgpt}, Seal-Tools \cite{wu2024seal}, and UltraTool \cite{huang2024planningcreationusagebenchmarking}. Despite covering a range of tool-use tasks, existing benchmarks have key limitations: (1) tool descriptions lack high-quality tool documentation; (2) missing parameters or type hints in ground truth; (3) insufficient count of tools in toolset to correctly evaluate the retrieval system. For our study, we selected Seal-Tools, BFCL, and UltraTool as primary datasets. Seal-Tools \cite{wu2024seal} includes out-of-domain test examples that call single and multiple tools over a very large toolset. BFCL \cite{patil2024gorilla} has a simple single-turn tool calling dataset. It contains a one-to-one query to tool mapping that allows us to test our tool merging strategy efficiently. UltraTool \cite{huang2024planningcreationusagebenchmarking} is specifically designed to improve an LLM's tool utilization within real-world scenarios. Details on the benchmark can be found in Table \ref{tab:datasets}.

\textbf{Evaluation Metrics.} 
We follow prior work \cite{lumer2024toolshed, wu2024seal} and use standard tool selection and retrieval-based metrics. Correct Selection Rate (CSR), adapted from MetaTool \cite{huang2023metatool}, measures the percentage of queries for which the predicted toolset exactly matches the ground-truth set. CSR@$k$ specifies the Correct Selection Rate using the top-$k$ tools retrieved from ToolScopeRetriever, as shown in Equation~\ref{eq:csr}:
\begin{equation}
\mathrm{CSR@}k = \frac{1}{|\mathcal{Q}|} \sum_{q \in \mathcal{Q}} \mathbb{I}[\hat{T}_q^{(k)} = T_q^*]
\label{eq:csr}
\end{equation}
Here, $\mathcal{Q}$ is the set of all evaluation queries, $T_q^*$ is the ground-truth toolset for query $q$, $\hat{T}_q^{(k)}$ is the predicted top-$k$ toolset, and $\mathbb{I}[\cdot]$ is the indicator function.

Recall@$k$ quantifies the proportion of ground-truth tools correctly retrieved among the top-$k$ predictions as shown in Equation~\ref{eq:recall}.
\begin{equation}
\mathrm{Recall@}k = \frac{1}{|\mathcal{Q}|} \sum_{q=1}^{|\mathcal{Q}|} \frac{|T_q^k \cap T_q^*|}{|T_q^*|}
\label{eq:recall}
\end{equation}

\textbf{Baselines.} We evaluate two retrieval baselines in all main CSR experiments: BM25 \cite{robertson2009probabilistic}, which ranks query–tool pairs by TF-IDF term overlap. Dense Embeddings \cite{qu2025tool}, which encodes queries and tool descriptions into a shared vector space and ranks them by cosine similarity. Both engines are evaluated on the original, unmerged tool sets, allowing any performance gap to be attributed to ToolScope. In multi-tool setting (e.g: UltraTool), we perform query decomposition and score normalization to select the top $k$ tools.

For completeness, we also include the additional baselines DPR \cite{wu2024seal} and ToolShed \cite{lumer2024toolshed} alongside BM25 and Dense Embeddings; their Seal-Tools retrieval scores appear in Table \ref{tab:recall_k_selected_configs} in Appendix \ref{app:D}.

\textbf{Implementation.} 
\\\textbf{ToolScopeMerger:} Each tool is embedded using \texttt{thenlper/gte-large} and indexed in FAISS (Facebook AI Similarity Search Library) \cite{douze2024faiss}. 
Tool pairs with cosine similarity above 0.82 are selected as merge candidates, a threshold determined 
via sensitivity analysis on Seal-Tools (Fig.~\ref{fig:sensitivity_plot}; Sec.~\ref{sec:discussion}). 
GPT-4o performs relationship classification, Auto-Correction, and toolset updating.
\textbf{ToolScopeRetriever:} We evaluate on the full toolset to reflect production settings, 
where the agent does not rely on pre-filtered tools (unlike UltraTool \cite{huang2024planningcreationusagebenchmarking}). 
Following results in Sec.~\ref{sec:discussion}, we use a dense-only retriever ($\alpha{=}1$) and rerank the 
top-50 candidates with a cross-encoder using min–max normalization. 
\textbf{Main Agent:} Tool selection are generated using 3 state-of-the-art models: GPT-4o \cite{openai2024gpt4o}, Cohere-Command-R-08-2024 \cite{cohere2024commandrefresh}, and LLaMA-3.3-70B \cite{llama2024llama3_3}.
\textbf{Result Analysis:} GPT-4o was used as LLM-as-a-Judge \cite{zheng2023judging} for grading documentation quality in Section~\ref{sec:discussion}.
\\ \textbf{Full implementation details and hyperparameters: }Full pseudo-code, prompts, hyperparameters, and hardware configurations are provided in Appendices~\ref{app:G}, \ref{app:H} and \ref{app:hyperparams}.

\subsection{Experimental Results}
We present the main results for LLM agent tool selection in Table~\ref{tab:tool_selection_models_first}, with full results provided in Appendix \ref{app:C}. Based on the results, we have the following observations: 

\textbf{ToolScope achieves state-of-the-art CSR.} 
We observe in Table~\ref{tab:tool_selection_models_first} that ToolScope, with or without Auto-Correction, achieves the highest CSR@$k$ on Seal-Tools, BFCL, and UltraTool across all $k$ values and LLMs including GPT\mbox{-}4o, Llama\mbox{-}3.3\mbox{-}70B, and Command\mbox{-}R. Gains are large on challenging datasets such as Seal-Tools (+34.6\%) and UltraTool (+38.6\%), and remain consistent on BFCL (+8.8\%). These results highlight ToolScope's strong generalization capability across different LLMs. Among the models evaluated, GPT-4o exhibits the highest CSR, indicating that ToolScope can benefit from continuous advancements in foundation models. Notably, ToolScopeMerger is not tied to GPT-4o, as experiments with open-source alternatives (LLaMA 3.3 70B and LLaMA 3.1 8B) result in comparable performance (see Appendix~\ref{tab:llm-ablation}).

\textbf{Robustness across $k$.} Gains are robust at all evaluated $k$. On Seal-Tools, the performance gap widens as $k$ increases, consistent with the hypothesis that the original toolset contains higher rates of overlap; the merged toolset avoids confusing the retriever with overlapped tools, preserving precision at larger $k$. On BFCL, improvements remain stable across $k$, and on UltraTool, ToolScope continues to outperform the baselines, showing scalability to larger numbers of retrieved tools passed to the LLM agent.

\textbf{Auto-Correction enhances CSR across datasets.} Auto-Correction is a lightweight LLM-based module from ToolScopeMerger that audits and refines merge decisions after graph-based consolidation. Across models, it consistently improves CSR. For example, it yields gains of 1.5--2.9\% on Seal-Tools and 2.9--7.9\% on UltraTool, where redundant or ambiguous tools are more common. Improvements on BFCL are modest for GPT-4o (+2.6\%) and LlaMa-3.3-70b (+5.0\%), but we observe a slight regression with command-R (–1.2\%), likely due to over-correction when the baseline coverage is already strong. These results suggest that Auto-Correction is most beneficial in complex or fast-evolving toolsets and can be deployed with guardrails (e.g., confidence thresholds or fallback merges) to prevent rare regressions. Appendix~\ref{app:J} illustrates this process with two case studies: one highlighting a successful merge and the other a rejected merge. These show the module’s ability to correct errors. 

\textbf{ToolScopeMerger preserves functionality while maintaining high merge reliability.} 
Using Tool-Call Coverage Rate (TCCR) and Unique Capability Coverage (UCC) (Appendix~\ref{app:functionality_preservation}), we observe 82--95\% call-level coverage and 80--96\% capability-level coverage across datasets (Table~\ref{tab:functionality_preservation}), indicating that most tool functionality is retained after merging. Preservation remains high even for infrequent capabilities ($\leq 3$ occurrences), with 82.0\% (BFCL), 95.8\% (Seal-Tools), and 86.4\% (UltraTool) retained. To assess merge quality, we conduct human evaluation on 48 BFCL tool clusters, showing 95.4\% correct merge decisions from ToolScopeMerger. In addition, the Auto-Correction module achieves 94.4\% F1 (95.5\% precision, 93.3\% recall), further supporting the reliability of the merging process (Appendix~\ref{app:human_eval_merges}).

\textbf{Preliminary end-to-end evaluation.}
We further conduct a small-scale end-to-end evaluation (20 queries) to assess whether improvements in tool selection translate to downstream task success. In a full tool-calling setting, ToolScope achieves 80\% final answer accuracy, substantially outperforming BM25 (30\%) and dense retrieval (40\%) (Appendix~\ref{tab:e2e}). These results provide initial evidence that gains in CSR can translate to improved end-to-end system performance.

\subsection{Ablation Study}
Table~\ref{tab:csr_checkmarks} evaluates the contributions of the three core components of ToolScope: the Reranker from ToolScopeRetriever, the Merger from ToolScopeMerger, and the Auto-Correction LLM module from ToolScopeMerger. Each dataset is assessed under different combinations of these modules. We also retain the base retrieval setup, including query decomposition, dense embedding, and score normalization. This serves as a stable retrieval foundation for reranking and merging decisions.

\textbf{ToolScopeMerger provides the strongest gains.} 
With Reranker off, enabling Merger boosts CSR by $22.0\%$ on Seal-Tools, $5.0\%$ on BFCL, and $7.0\%$ on UltraTool. The larger improvements on Seal-Tools and UltraTool stem from their higher semantic redundancy, where merging prevents confusing overlaps in the retrieved pool. Beyond accuracy, ToolScopeMerger also reduces raw toolset size: $-14.0\%$ on BFCL, $-2.1\%$ on Seal-Tools, and $-25.3\%$ on UltraTool (Table~\ref{tab:toolset_merge_comparison}, Appendix \ref{app:D}), providing a cleaner foundation for retrieval.

\textbf{Auto-Correction further improves performance by refining merge errors}, as seen in Table \ref{tab:tool_selection_models_first}. When added on top of Reranker + Merger, it raises CSR by $2.9\%$ on Seal-Tools, $5.0\%$ on BFCL, 
and $7.9\%$ on UltraTool (Table~\ref{tab:tool_selection_models_first}). 
These gains indicate that even after graph-based consolidation, some false merges remain and can 
be efficiently corrected by a lightweight LLM audit.

\textbf{Reranker helps most on large, noisy toolsets.} 
Its gains are $+1.3\%$ on Seal-Tools and $+0.9\%$ on UltraTool, but negligible ($0.0\%$) on BFCL. 
This reflects dataset differences: BFCL is a single-tool benchmark where retrieval already isolates the correct tool, while Seal-Tools and UltraTool contain larger, noisier multi-tool corpora where reranking helps disambiguate overlapping candidates. We therefore keep Reranker in the pipeline, as it consistently adds value on more challenging datasets.

In summary, ToolScopeMerger delivers the dominant accuracy gain, Auto-Correction provides consistent refinements, and Reranker offers selective improvements. These components help ToolScope achieve strong and consistent performance across all three benchmarks. 

\begin{table}[t]
\centering
\small
\begin{tabular}{lccc c}
\toprule
\textbf{Dataset} & \textbf{Reranker} & \textbf{Merger} & \textbf{AutoCorrect} & \textbf{CSR} \\
\midrule
\multirow{3}{*}{Seal-Tools}
  & \cmark & \cmark & \cmark & 0.931 \\
  & \cmark & \xmark & \xmark & 0.912 \\
  & \xmark & \cmark & \cmark & 0.918 \\
  & \xmark & \xmark & \xmark & 0.694 \\
\midrule
\multirow{3}{*}{BFCL}
  & \cmark & \cmark & \cmark & 0.938 \\
  & \cmark & \xmark & \xmark & 0.888 \\
  & \xmark & \cmark & \cmark & 0.938 \\
  & \xmark & \xmark & \xmark & 0.888 \\
\midrule
\multirow{4}{*}{UltraTool}
  & \cmark & \cmark & \cmark & 0.752 \\
  & \cmark & \xmark & \xmark & 0.622 \\
  & \xmark & \cmark & \cmark & 0.743 \\
  & \xmark & \xmark & \xmark & 0.673 \\
\bottomrule
\end{tabular}
\caption{Configuration usage (\cmark/\xmark) for each dataset and corresponding CSR score. BFCL is evaluated at $k=5$, Seal-Tools is evaluated at $k=30$, and UltraTool is evaluated at $k=30$. All results are generated using GPT-4o.}
\label{tab:csr_checkmarks}
\end{table}

\subsection{Discussion}
The findings of this study provide important insights into the following points.

\textbf{ToolScopeMerger effectively reduces functional overlap.}  
ToolScopeMerger removes semantically similar tools, improving the distribution in embedding space. We also evaluated the impact using the silhouette coefficient \cite{rousseeuw1987silhouettes}, where lower values indicate less semantic overlap. In addition, we used t-SNE plots \cite{van2008visualizing} to visualize the tool semantic distribution. As shown in Figure \ref{fig:tsne_vis} (details in Appendix \ref{app:E2}), Seal-Tools, UltraTool and BFCL all exhibit sparser distributions. As seen in Appendix \ref{app:E1}, all datasets also have lower silhouette scores after merging with varying cluster sizes, confirming that ToolScopeMerger reduces semantic redundancy.

\begin{figure}[t] 
  \centering
  \includegraphics[width=\columnwidth]{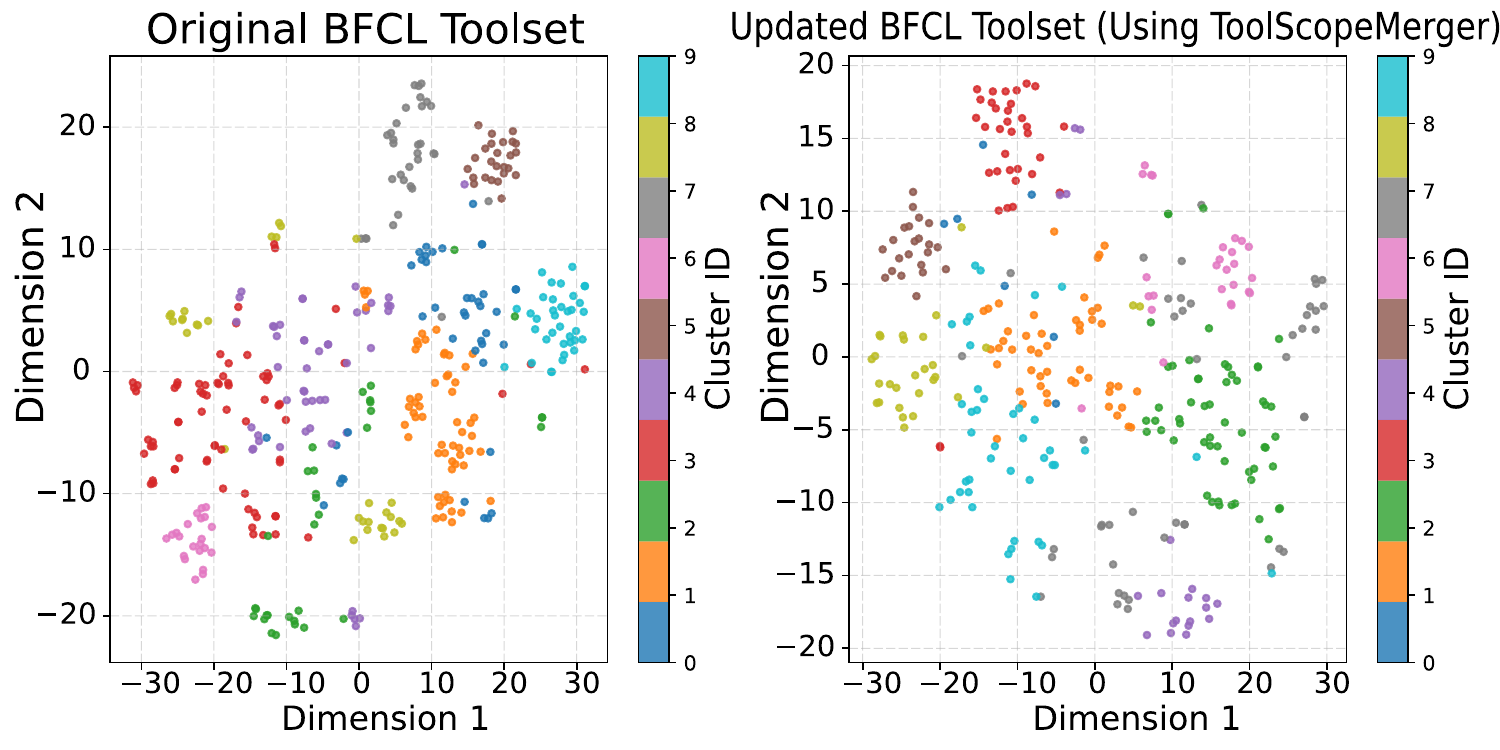}
  \caption{T-SNE visualization of original BFCL tool embedding and merged BFCL tool embedding}
  \label{fig:tsne_vis}
\end{figure}

\begin{figure}[t]
  \centering
  \includegraphics[width=0.8\columnwidth]{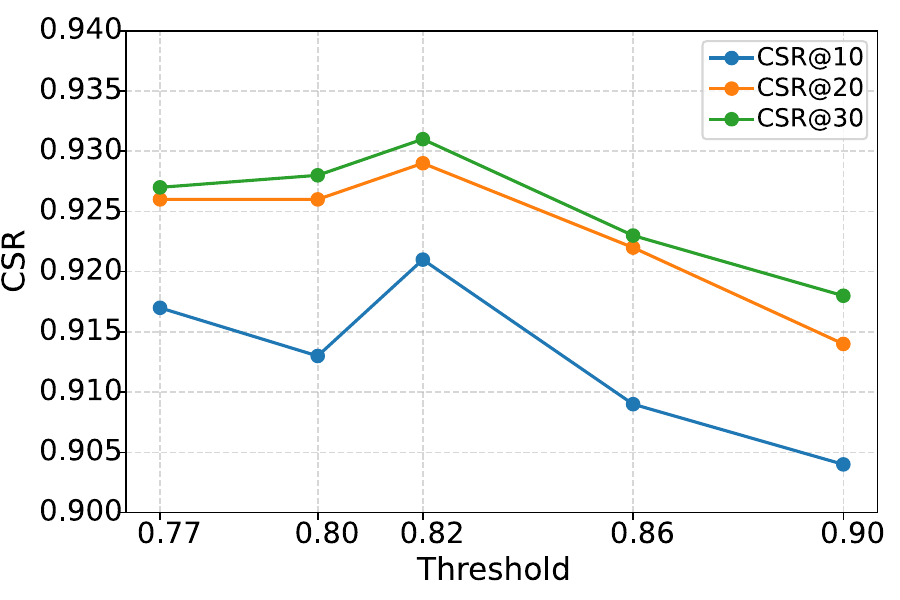}
  \caption{Sensitivity analysis of ToolScopeMerger on Seal-Tools using different cosine similarity filtering thresholds.}
  \label{fig:sensitivity_plot}
\end{figure}

\textbf{ToolScopeMerger is robust to threshold tuning.}  
As shown in Figure~\ref{fig:sensitivity_plot}, we conducted a sensitivity study on the cosine similarity threshold in ToolScopeMerger, using the most heterogeneous corpus, Seal-Tools, and sweeping from 0.77 up to 0.90. CSR remains stable in the 0.77 to 0.82 band, with the three lowest thresholds differing by less than 0.5\%. CSR reaches its maximum at 0.82. For thresholds above 0.86, CSR declines, indicating that overly conservative merging can discard useful semantically similar tool pairs. We therefore fix the cosine similarity threshold at 0.82 for all experiments. The consistently high CSR values across thresholds from 0.77 to 0.90 indicate that ToolScopeMerger is robust to a broad range of reasonable settings.

\textbf{ToolScopeMerger ensures merge validation without the need for manual intervention.}
We eliminate manual validation by introducing Auto-Correction, an LLM-based module that audits and refines merge decisions (Section~\ref{sec:merger}). It corrects merge errors and improves CSR by 1.5--2.9\% on Seal-Tools and 2.9--7.9\% on UltraTool, where tool redundancy is high. Gains on BFCL are moderate: 2.6\% (GPT-4o) and 5.0\% (LlaMa-3.3-70b). Beyond accuracy, Auto-Correction reduces the need of manual inspection and enables developers to review and improve tool definitions as needed.

\begin{figure}[t]
  \centering
  \includegraphics[width=\columnwidth]{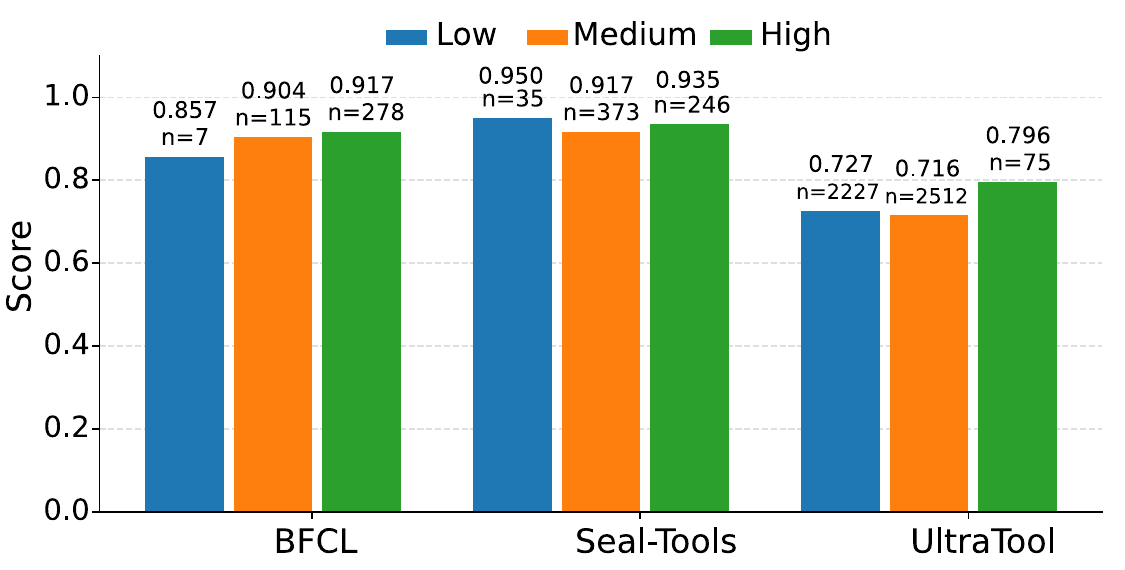}
  \caption{CSR scores for BFCL, Seal-Tools, and UltraTool by tool quality (Low/Medium/High). $n$ denotes the number of evaluation queries for each quality level used to compute the CSR.}
  \label{fig:quality_barplot}
\end{figure}

\textbf{Robustness Analysis of Documentation Quality.}
We evaluate whether ToolScope depends on high-quality documentation by scoring each tool’s name, signature, and description on a 1–5 scale using GPT-4o as an LLM judge (Appendix \ref{app:H5}). Tools are grouped into Low (1–2), Medium (3), and High (4–5) quality bands, and CSR is analyzed within each band. As shown in Figure~\ref{fig:quality_barplot}, ToolScope maintains strong CSR even with low-quality documentation: 85.7\% (BFCL), 95.0\% (Seal-Tools), and 72.7\% (UltraTool). Although Seal-Tools and UltraTool contain mostly medium or low-quality tools (79\% and 96\%, respectively), ToolScope remains robust to imperfect descriptions. Human validation of GPT-4o’s scoring confirms strong alignment with human judgment (Appendix \ref{app:F2}), supporting the reliability of this evaluation. These results show that ToolScope remains effective even with brief or imperfect documentation, as long as each tool includes a name, signature, and short description, conditions typical of real-world toolsets.

\textbf{ToolScope improves retrieval performance.}  
ToolScopeMerger enhances retrieval accuracy by reducing semantic overlap in the toolset (Table \ref{tab:recall_k_selected_configs} in Appendix \ref{app:D}). Recall@10 increases significantly, from 0.550 to 0.935 on Seal-Tools and from 0.945 to 0.985 on BFCL. ToolScopeRetriever uses a hybrid strategy combining sparse (BM25) and dense embeddings, with a weighting parameter $\alpha$ controlling their balance. Our $\alpha$ tuning experiment (Figure \ref{fig:alpha_recall_sealtools}, Appendix \ref{app:D}) shows that retrieval@$k$ peaks at $\alpha=1$, where dense-only retrieval performs best. This configuration outperforms baselines across all $k$ on BFCL and at $k=5$ on Seal-Tools.

\textbf{Reranking improves CSR at small retrieval $k$.}  
As shown in Figure \ref{fig:single_tool_csr} from Appendix \ref{app:D}, enabling the reranker significantly improves CSR at smaller retrieval sizes ($k=5$ or $10$). As $k$ increases, the performance gap between reranked and non-reranked configurations narrows, suggesting diminishing returns. Overall, $k=5$ or $k=10$ with reranking provides a good balance between performance and retrieval efficiency.

\textbf{ToolScopeRetriever reduces context length.} 
ToolScopeRetriever drastically 
shrinks the number of tokens passed to the agent by ranking and selecting only the top-$k$ relevant tools per query. As shown in Table~\ref{tab:context_reduction} in Appendix \ref{app:context_length}, 
average total context length is reduced from 32{,}563 to 469 tokens on BFCL ($98.6\%$ reduction), 
from 292{,}107 to 317 on Seal-Tools ($99.9\%$), and from 136{,}352 to 2{,}076 on UltraTool 
($98.5\%$). These reductions make the approach efficient and scalable even for very large toolsets 
and strict prompt budgets.

\textbf{Tool overlap remains a limitation of open-source benchmarks.}  
While Toolshed \cite{lumer2024toolshed} notes some overlap in Seal-Tools \cite{wu2024seal}, we identified many semantically redundant tools that compromise retrieval and selection accuracy without a solution like ToolScopeMerger. Similar issues appear in BFCL \cite{patil2024gorilla} and UltraTool \cite{huang2024planningcreationusagebenchmarking}. Such overlap introduces ambiguity that degrades retrieval metrics and causes selection errors. It also limits the reliability of benchmarking research in tool learning. Benchmark creators should enforce clearer distinctions or merge overlapping tools to enable more consistent evaluation. ToolScope offers a practical remedy for both production and research scenarios involving noisy toolsets.
\label{sec:discussion}

\section{Conclusion}
In this paper, we present \textbf{ToolScope}, a two-part framework aimed at addressing key challenges in LLM agent tool selection: tool overlap and limitations of long context length. \textbf{ToolScopeMerger} consolidates overlapping tools using an automated framework; its Auto-Correction module automatically audits and fixes merges, reducing semantic redundancy in large toolsets. \textbf{ToolScopeRetriever} improves retrieval accuracy through query decomposition, hybrid retrieval, reranker and robust score normalization for tool retrieval. Together, our results demonstrate that ToolScope significantly improves tool selection accuracy across standard benchmarks, achieving gains of 34.6\% on Seal-Tools, 38.6\% on UltraTool, and 8.8\% on BFCL over strong baselines. We also identify a persistent tool overlap issue in currently available open-source public benchmarks, which can be improved using ToolScope. Overall, ToolScope provides a scalable solution for improving LLM-agent tool selection in real-world settings. 

\section*{Limitations}
Our retrieval currently relies on tool names, signatures, and descriptions. 
Enriching tools with metadata such as typical use cases, domains, and input/output schemas could 
further strengthen ranking, particularly in very large toolsets. Complementary strategies like 
multi-index retrieval may also improve scalability. Finally, evaluating with reasoning-oriented 
models could influence the effectiveness of Auto-Correction in ToolScopeMerger and the quality 
of tool selection.

\section*{Ethical Considerations}

While ToolScope improves LLM agent performance in tool selection tasks, it is not yet suitable for deployment in environments where errors in tool selection could result in significant harm or consequences such as those in medical, legal, or financial fields. ToolScope relies on LLM generated tool merging and retrieval mechanisms which are inherently probabilistic. As with many LLM solutions, this is subject to hallucination, bias, and misclassifications of overlapped tools. 

\bibliography{custom}            
\nocite{*}                       

\input{appendix}

\end{document}

%% file: appendix.tex
\title{Appendix}

\maketitle

\appendix
\makeatletter
\renewcommand\section{\@startsection{section}{1}{\z@}%
  {2.0ex plus 1ex minus .2ex}%
  {1.0ex plus .2ex}%
  {\normalfont\Large\bfseries\raggedright}}
\makeatother

\section{Future Work}
We expect additional improvements in our future work. First, we aim to adopt a more extensible retrieval method such as a multi-index framework that which enhance scalability and relevance in large toolsets. Second, we plan to enrich the tool metadata with descriptive metadata (e.g: common use case descriptions, domain and user intents) and structural metadata (e.g: input and output schemas) that could provide stronger retrieval context and improve the retrieval accuracy. Finally, we want to expand the scope of the evaluation to analyze ToolScope’s impact on tool calling and response generation across multiple datasets. This would help to establish a clear picture of holistic improvements to LLM agent tool learning.
\label{app:B}



\section{Full CSR Scores for Baseline and ToolScope}
We report the full Correct Selection Rate (CSR@$k$) results for tool selection across BFCL, Seal-Tools, and UltraTool in Table~\ref{tab:csr_grouped_by_dataset} on different $k$ (where $k$ refers to the top $k$ tools retrieved for LLM Agent tool selection). With or without Auto-Correction, ToolScope consistently outperforms BM25  and Dense baselines, resulting in large improvements on Seal-Tools (+34.6\%) and UltraTool (+38.6\%), and stable improvements on BFCL (+8.8\%). These results demonstrate its strong cross model generalization capability, with GPT-4o as the generation model achieving the highest CSR.
\label{app:C}

\begin{table*}[!htbp]
\centering
\small

\begin{tabularx}{\textwidth}{
  >{\raggedright\arraybackslash}l
  >{\raggedright\arraybackslash}l
  >{\raggedright\arraybackslash}l
  *{6}{>{\centering\arraybackslash}X}
}
\toprule
\textbf{Dataset} & \textbf{Model} & \textbf{Config} & CSR@$5$ & CSR@$10$ & CSR@$15$ & CSR@$20$ & CSR@$25$ & CSR@$30$ \\
\midrule
\multirow{12}{*}{Seal-Tools}
  & \multirow{4}{*}{GPT-4o} & BM25 & 0.544 & 0.593 & 0.646 & 0.666 & 0.702 & 0.727 \\
  &  & Dense & 0.548 & 0.603 & 0.631 & 0.657 & 0.676 & 0.694 \\
  &  & ToolScope & 0.875 & 0.896 & 0.905 & 0.908 & 0.902 & 0.908 \\
  &  & ToolScope+AC & \textbf{0.890} & \textbf{0.921} & \textbf{0.924} & \textbf{0.929} & \textbf{0.931} & \textbf{0.931} \\
\cmidrule(lr){3-9}
  & \multirow{4}{*}{LLaMA-3.3-70B} & BM25 & 0.527 & 0.574 & 0.618 & 0.653 & 0.692 & 0.709 \\
  &  & Dense & 0.531 & 0.584 & 0.611 & 0.637 & 0.657 & 0.663 \\
  &  & ToolScope & 0.883 & 0.900 & 0.910 & 0.910 & 0.910 & 0.910 \\
  &  & ToolScope+AC & \textbf{0.889} & \textbf{0.916} & \textbf{0.922} & \textbf{0.924} & \textbf{0.926} & \textbf{0.928} \\
\cmidrule(lr){3-9}
  & \multirow{4}{*}{Command-R-08-2024} & BM25 & 0.463 & 0.512 & 0.551 & 0.573 & 0.593 & 0.608 \\
  &  & Dense & 0.523 & 0.569 & 0.603 & 0.621 & 0.647 & 0.666 \\
  &  & ToolScope & 0.873 & 0.888 & 0.889 & 0.891 & 0.890 & 0.887 \\
  &  & ToolScope+AC & \textbf{0.878} & \textbf{0.900} & \textbf{0.902} & \textbf{0.904} & \textbf{0.908} & \textbf{0.905} \\
\bottomrule
\end{tabularx}

\begin{tabularx}{\textwidth}{
  >{\raggedright\arraybackslash}l  
  >{\raggedright\arraybackslash}l  
  >{\raggedright\arraybackslash}l  
  *{4}{>{\centering\arraybackslash}X}  
}
\toprule
\textbf{Dataset} & \textbf{Model} & \textbf{Config} & CSR@$1$ & CSR@$5$ & CSR@$10$ & CSR@$15$ \\
\midrule
\multirow{12}{*}{BFCL}
  & \multirow{4}{*}{GPT-4o} & BM25 & 0.695 & 0.850 & 0.875 & 0.870 \\
  &  & Dense & 0.780 & 0.888 & 0.900 & 0.915 \\
  &  & ToolScope & \textbf{0.878} & 0.912 & 0.915 & 0.915 \\
  &  & ToolScope+AC & 0.860 & \textbf{0.938} & \textbf{0.935} & \textbf{0.940} \\
\cmidrule(lr){3-7}
  & \multirow{4}{*}{LLaMA-3.3-70B} & BM25 & 0.692 & 0.832 & 0.848 & 0.865 \\
  &  & Dense & 0.782 & 0.840 & 0.845 & 0.862 \\
  &  & ToolScope & \textbf{0.872} & 0.888 & 0.892 & 0.882 \\
  &  & ToolScope+AC & 0.858 & \textbf{0.935} & \textbf{0.942} & \textbf{0.942} \\
\cmidrule(lr){3-7}
  & \multirow{4}{*}{Command-R-08-2024} & BM25 & 0.695 & 0.785 & 0.805 & 0.807 \\
  &  & Dense & 0.785 & 0.802 & 0.825 & 0.822 \\
  &  & ToolScope & \textbf{0.872} & 0.902 & \textbf{0.912} & \textbf{0.920} \\
  &  & ToolScope+AC & 0.858 & \textbf{0.908} & 0.900 & 0.900 \\
\bottomrule
\end{tabularx}

\begin{tabularx}{\textwidth}{
  >{\raggedright\arraybackslash}l
  >{\raggedright\arraybackslash}l
  >{\raggedright\arraybackslash}l
  *{2}{>{\centering\arraybackslash}X}
}
\toprule
\textbf{Dataset} & \textbf{Model} & \textbf{Config} & CSR@$20$ & CSR@$30$ \\
\midrule
\multirow{12}{*}{UltraTool}
  & \multirow{4}{*}{GPT-4o} & BM25 & 0.333 & 0.366 \\
  &  & Dense & 0.647 & 0.673 \\
  &  & ToolScope & 0.691 & 0.723 \\
  &  & ToolScope+AC & \textbf{0.702} & \textbf{0.752} \\
\cmidrule(lr){3-5}
  & \multirow{4}{*}{LLaMA-3.3-70B} & BM25 & 0.330 & 0.364 \\
  &  & Dense & 0.649 & 0.664 \\
  &  & ToolScope & 0.692 & 0.729 \\
  &  & ToolScope+AC & \textbf{0.710} & \textbf{0.759} \\
\cmidrule(lr){3-5}
  & \multirow{4}{*}{Command-R-08-2024} & BM25 & 0.338 & 0.368 \\
  &  & Dense & 0.648 & 0.669 \\
  &  & ToolScope & 0.676 & 0.702 \\
  &  & ToolScope+AC & \textbf{0.688} & \textbf{0.718} \\
\bottomrule
\end{tabularx}

\caption{Correct Selection Rate (CSR) of top $k$ tools across datasets (BFCL, Seal-Tools, UltraTool) and models. $k$ refers to the top $k$ tools retrieved for LLM Agent tool selection. Configurations compared: BM25, Dense, ToolScope, and ToolScope + AC (ToolScope with Auto-Correction). Seal-Tools is evaluated at $k\in\{5,10,15,20,25,30\}$, BFCL at $k\in\{1,5,10,15\}$,  and UltraTool at $k\in\{20,30\}$.}
\label{tab:csr_grouped_by_dataset}
\end{table*}

\section{Retrieval Performance for ToolScope}
We evaluate tool retrieval performance across both the Seal-Tools and BFCL benchmarks using Recall@$k$, which measures the proportion of ground-truth tools appearing in the top-$k$ retrieved candidates.
Table~\ref{tab:recall_k_selected_configs} summarizes the results for several representative configurations.
We observe that our solution achieves the highest recall on both benchmarks.
\label{app:D}

\begin{table}[H]
\centering
\small
\begin{threeparttable}
\begin{tabularx}{\linewidth}{
  >{\raggedright\arraybackslash}X             
  >{\centering\arraybackslash}p{1.0cm}        
  >{\centering\arraybackslash}p{1.0cm}        
  >{\centering\arraybackslash}p{1.0cm}        
}
\toprule
\textbf{Benchmark / Configuration} & \multicolumn{3}{c}{\textbf{Recall@$k$}} \\
\cmidrule(lr){2-4}
& \textbf{$k=1$} & \textbf{$k=5$} & \textbf{$k=10$} \\
\midrule
Seal-Tools / BM25  & - & 0.490 & 0.540 \\
Seal-Tools / Dense & - & 0.589 & 0.649 \\
Seal-Tools / DPR & - & 0.480 & 0.680 \\
Seal-Tools / ToolShed & - & 0.876 & \textbf{0.965} \\
Seal-Tools / ToolScope & - & \textbf{0.884} & 0.935 \\
\midrule
BFCL / BM25 & 0.693 & 0.913 & 0.945 \\
BFCL / Dense & 0.818 & 0.973 & 0.985 \\
BFCL / ToolScope & \textbf{0.880} & \textbf{0.973} & \textbf{0.985} \\
\bottomrule
\end{tabularx}
\end{threeparttable}
\caption{Recall@$k$ scores for selected retrieval configurations on Seal-Tools and BFCL benchmarks. ToolScope achieves the highest recall across both benchmarks for $k=5$ for Seal-Tools and all $k \in \{1,5,10\}$ for BFCL, and competitive recall score for $k=1$ for Seal-Tools.}
\label{tab:recall_k_selected_configs}
\end{table}

\begin{figure}[H]
  \centering
  \includegraphics[width=\columnwidth]{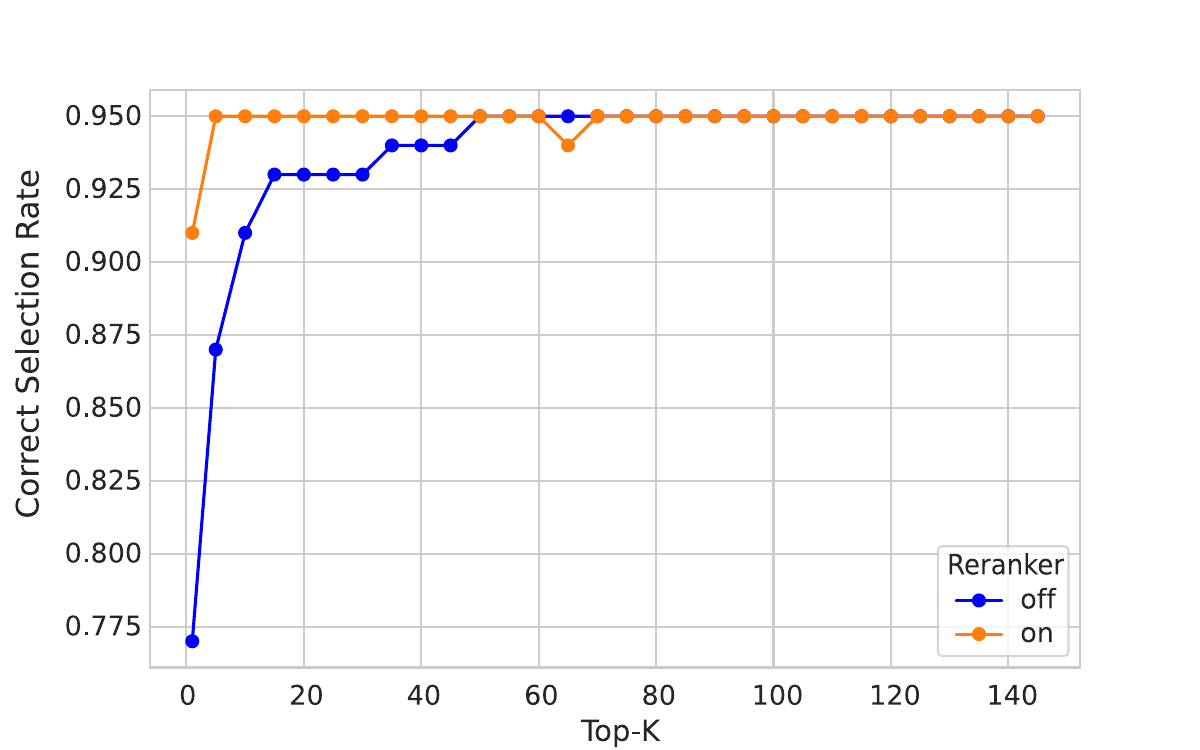}
  \caption{The CSR results (\%) of top $k$ tools with/without Reranker}
  \label{fig:single_tool_csr}
\end{figure}

Figure~\ref{fig:single_tool_csr} illustrates the Correct Selection Rate (CSR@$k$) with and without the Reranker. Across all datasets, ToolScope with Reranker consistently outperforms ToolScope without Reranker. As $k$ increases, the scores gradually converge, indicating that the benefit of reranking diminishes when enough tools are retrieved. This highlights the Reranker’s effectiveness in improving tool selection, particularly when fewer candidate tools are considered.

\begin{figure}[H]
  \centering
  \includegraphics[width=\columnwidth]{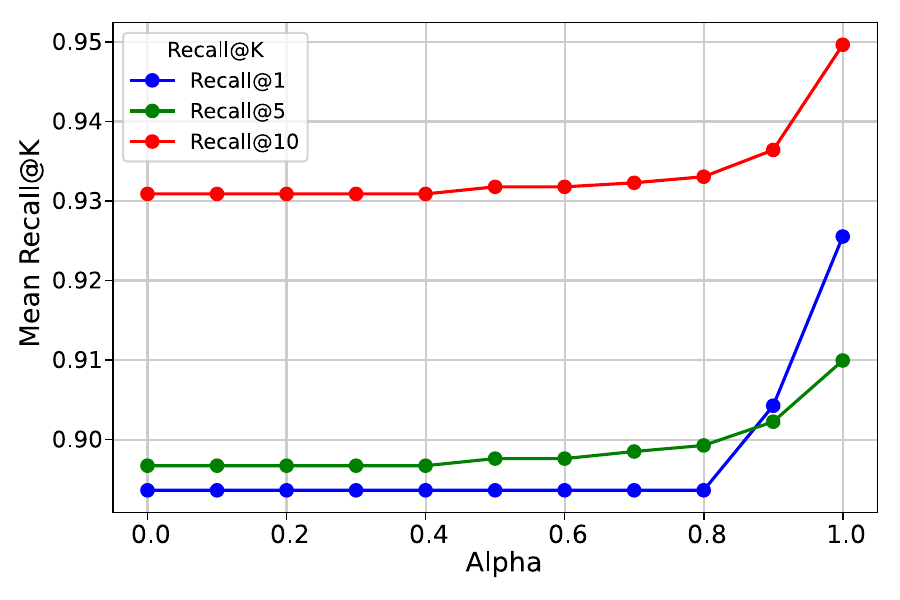}
  \caption{Recall@$k$ for different values of $\alpha$ on the Seal-Tools dataset.}
  \label{fig:alpha_recall_sealtools}
\end{figure}

We analyze the effect of the hybrid retrieval weight $\alpha$ on retrieval performance in Seal-Tools, as shown in Figure~\ref{fig:alpha_recall_sealtools}. $\alpha$ controls the weightage given to BM25 and Dense scores. Performance steadily increases with higher values of $\alpha$, peaking at $\alpha=1$, which shows that prioritizing the dense retriever results in the best recall@$k$ scores overall.

ToolScopeMerger reduces toolset redundancy across all benchmarks (Table~\ref{tab:toolset_merge_comparison}). Merging results in moderate reductions for BFCL (–14.0\%) and Seal-Tools (–2.1\%), and a substantial reduction for UltraTool (–25.3\%).

\begin{table}[H]
\centering
\small
\textbf{Toolset Size Comparison After Merging}

\resizebox{\linewidth}{!}{
\begin{tabular}{
lccc
}
\toprule
\textbf{Dataset} & \textbf{Original Size} & \textbf{Merged Size} & \textbf{\% Change} \\
\midrule
BFCL       & 400  & 344  & -14.0\% \\
Seal-Tools & 4076 & 3992 & -2.1\% \\
UltraTool  & 1885 & 1408 & -25.3\% \\
\bottomrule
\end{tabular}
}

\caption{Comparison of benchmark toolset sizes before and after merging with ToolScopeMerger.}
\label{tab:toolset_merge_comparison}
\end{table}

\section{ToolScopeMerger Results}
\label{app:E}

\subsection{Silhouette Scores on Seal-Tools, BFCL, UltraTool}

\begin{figure}[!htbp]
  \centering
  \includegraphics[width=\columnwidth]{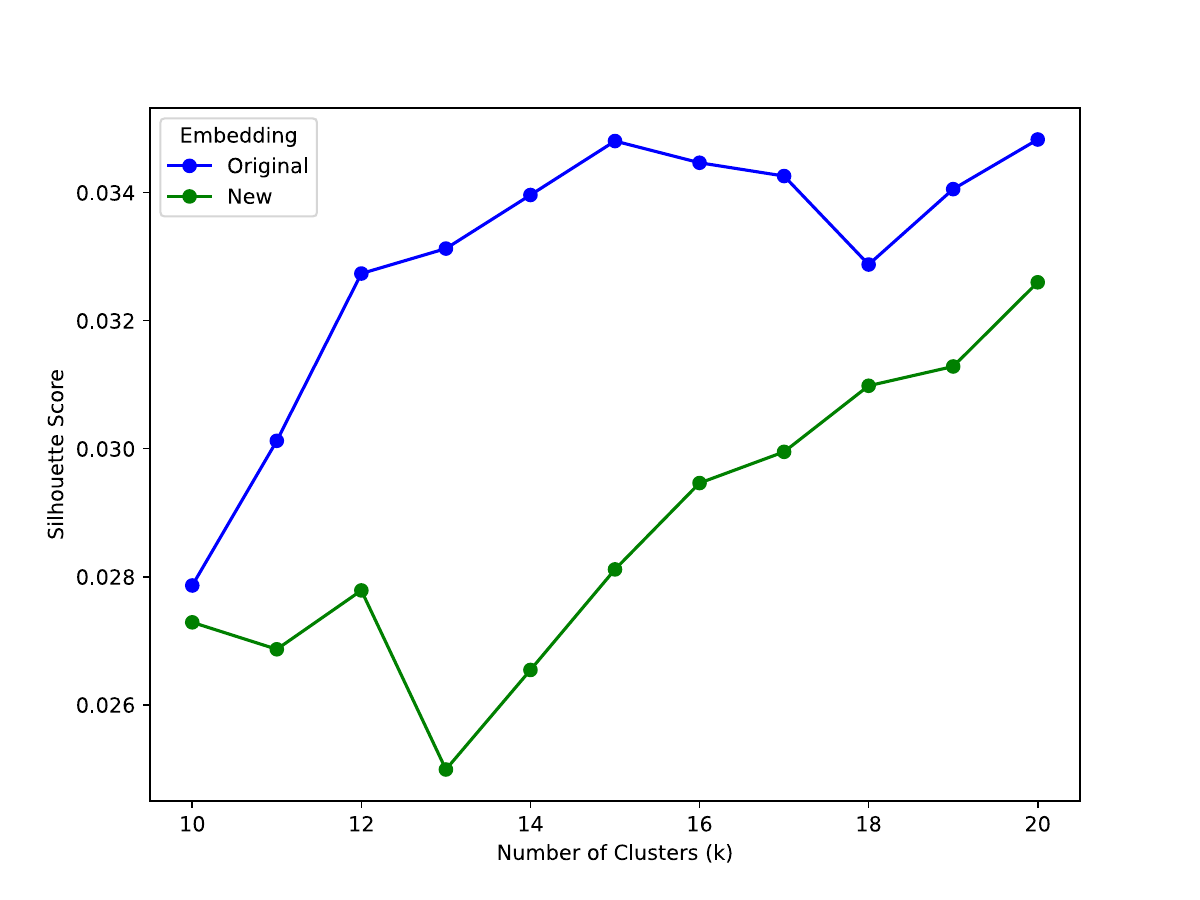}
  \caption{Seal-Tools Silhouette Scores}
  \label{fig:sealtool_silo_autocorrect}
\end{figure}

\begin{figure}[!htbp]
  \centering
  \includegraphics[width=\columnwidth]{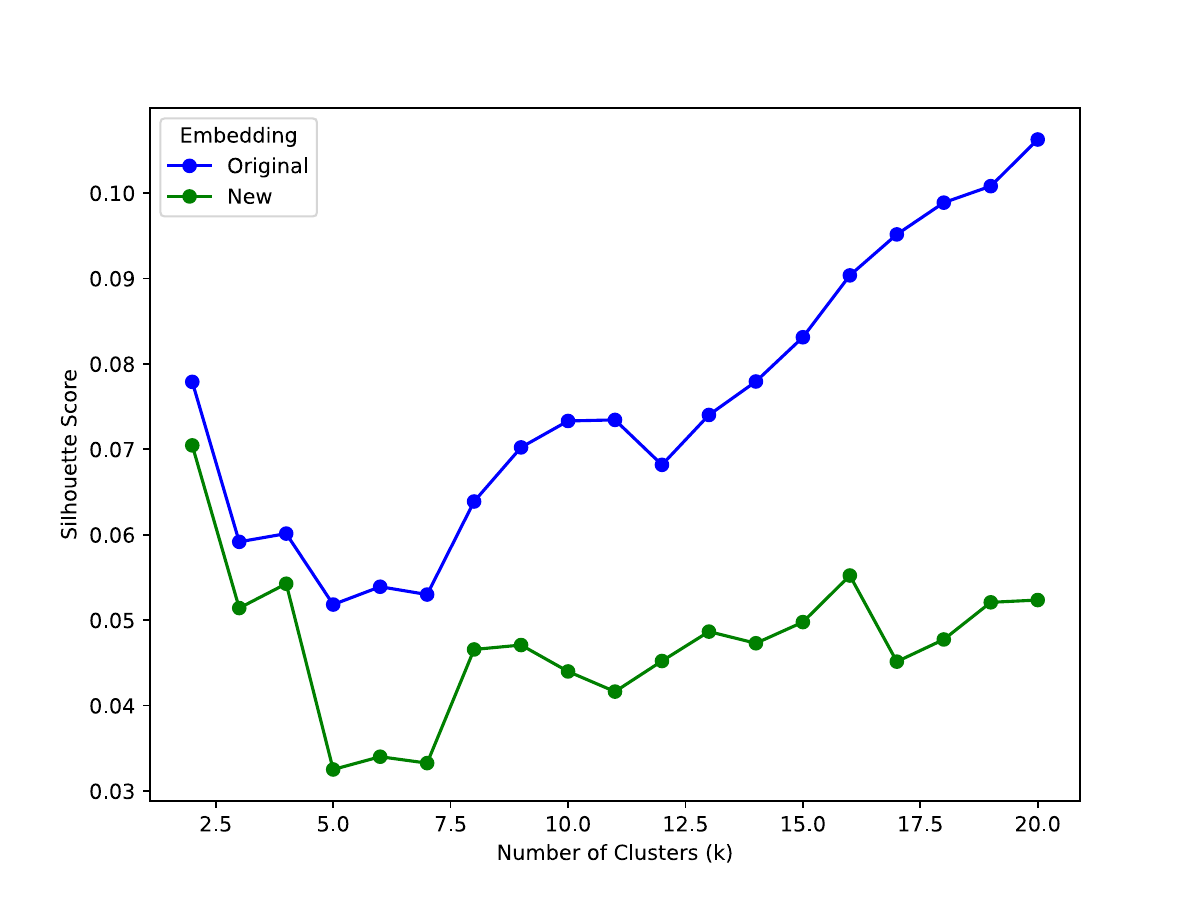}
  \caption{BFCL Silhouette Scores}
  \label{fig:bfcl_silo_autocorrect}
\end{figure}

\begin{figure}[!htbp]
  \centering
  \includegraphics[width=\columnwidth]{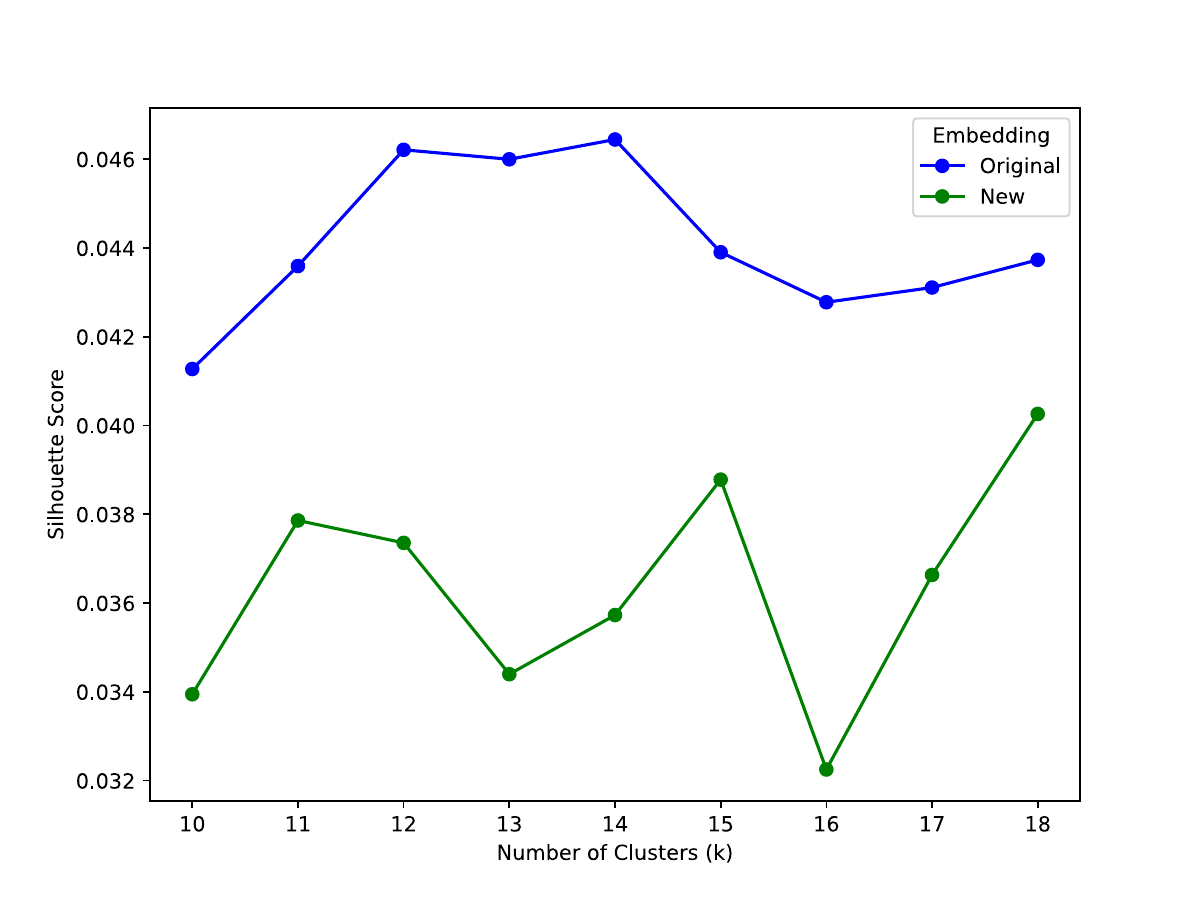}
  \caption{UltraTool Silhouette Scores}
  \label{fig:ultratool_silo_autocorrect}
\end{figure}

Figures~\ref{fig:sealtool_silo_autocorrect},~\ref{fig:bfcl_silo_autocorrect}, and~\ref{fig:ultratool_silo_autocorrect} show the silhouette scores for each of our benchmark datasets before and after applying ToolScopeMerger with Auto-Correction. Silhouette score is used to assess clustering quality, ranging from -1 to 1. A high silhouette score can reflect a high tool overlap across clusters, since tools that are overly similar will cluster tightly.

After applying our solution, the silhouette scores consistently decrease across different values of k, indicating a reduction in tool overlap. This suggests that ToolScopeMerger with Auto-Correction successfully merges redundant tools and increases the functional diversity across the toolset.
\label{app:E1}

\subsection{ToolScopeMerger T-SNE Plots on Seal-Tools, UltraTool}
\label{app:E2}

\begin{figure}[!htbp]
  \centering
  \includegraphics[width=\columnwidth]{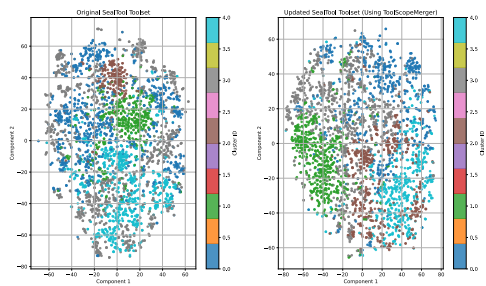}
  \caption{T-SNE visualization of original Seal-Tools tool embedding and merged and autocorrected Seal-Tools tool embedding}
  \label{fig:sealtool_merge_autocorrect}
\end{figure}

\begin{figure}[!htbp]
  \centering
  \includegraphics[width=\columnwidth]{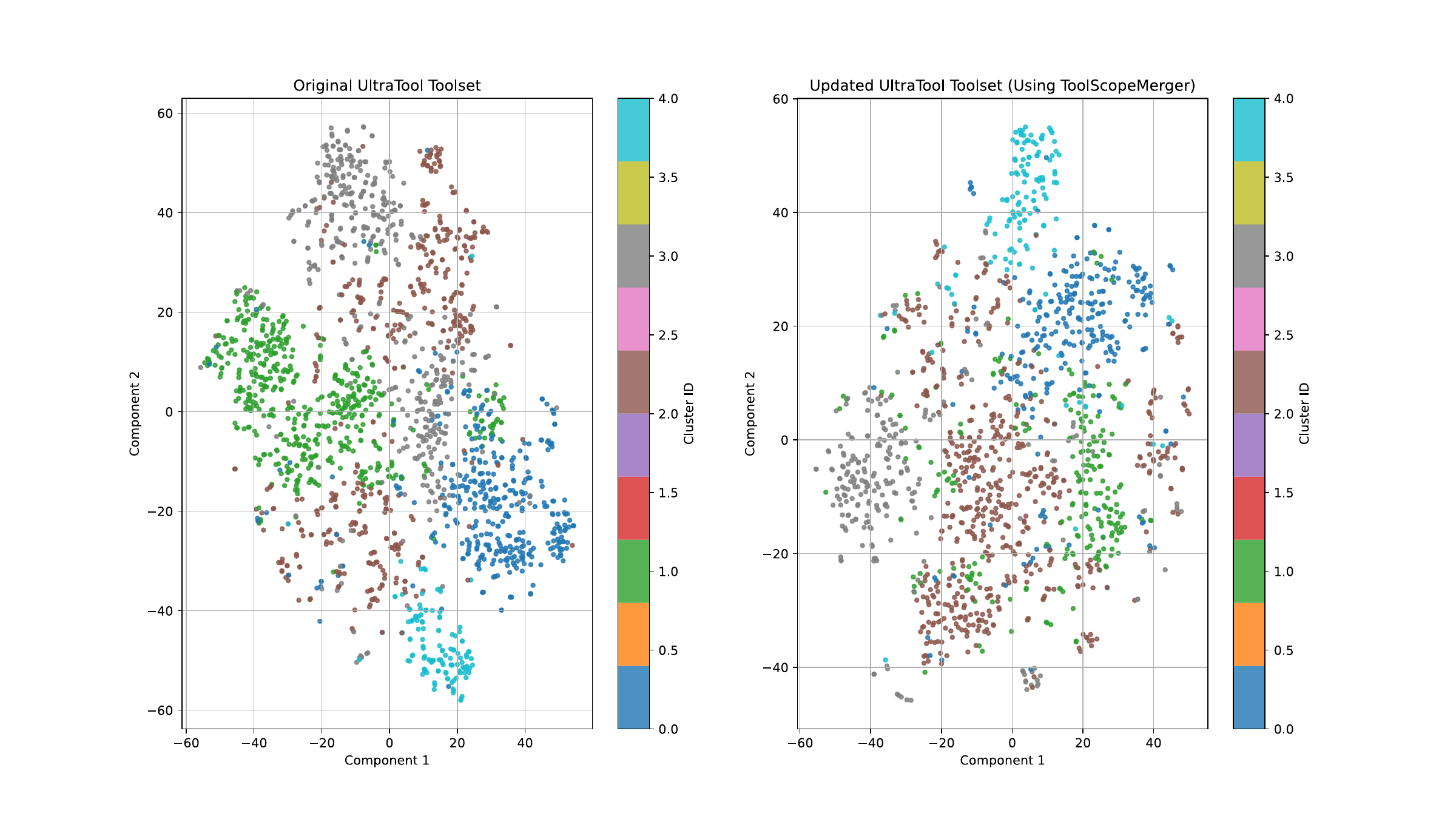}
  \caption{T-SNE visualization of original UltraTool tool embedding and merged and autocorrected UltraTool tool embedding}
  \label{fig:ultratool_merge_autocorrect}
\end{figure}

Figures~\ref{fig:sealtool_merge_autocorrect} and~\ref{fig:ultratool_merge_autocorrect} present T-SNE plots of the benchmark datasets before and after we apply ToolScopeMerger with Auto-Correction. T-SNE is a technique that allows us to visualize high-dimensional data in 2D space. After applying our solution, the plots show visibly reduced cluster overlap. This indicates that ToolScopeMerger effectively merges redundant tools and improves separation between tool groups. 

\section{Tool Documentation Quality Analysis}
\label{app:F}

\subsection{Tool Documentation Quality Overview by Dataset}
\begin{figure}[!htbp]
  \centering
  \includegraphics[width=\columnwidth]{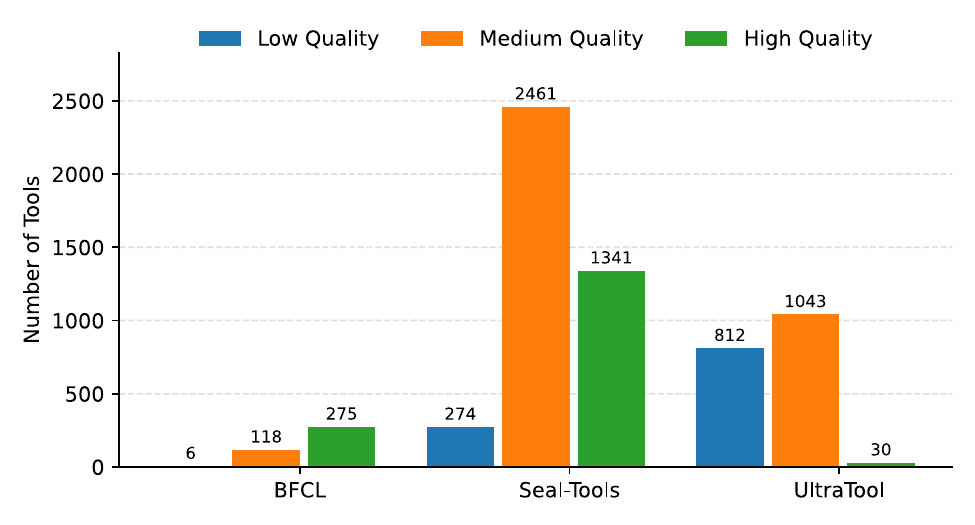}
  \caption{Tool Counts by Quality (Low, Medium, High) Across Benchmarks}
  \label{fig:tools_quality_counts}
\end{figure}

Figure~\ref{fig:tools_quality_counts} shows the tool quality distribution for each dataset, adding extra context to our robustness analysis on ToolScope. As seen in the result, Seal-Tools and UltraTool contain a substantial number of medium and low-quality tools, while BFCL is dominated by high-quality ones. Upon human verification, we have also confirmed the low quality tools from Seal-Tools and UltraTool still contain a basic structure of the tools: a name, signature, and a brief description.
\label{app:F1}

\subsection{Human Agreement and GPT-4o Alignment}
To assess the reliability of our documentation quality scores and validate the use of GPT-4o as an LLM judge, we conducted a small-scale human annotation study.
\paragraph{Human Annotation Protocol.}
3 independent annotators evaluated a sample of 30 tools drawn from Seal-Tools and UltraTool. Tools were stratified into three bins (Low, Medium, High) based on initial GPT-4o ratings, with 10 tools sampled per bin. Annotators rated each tool’s documentation using the same rubric provided to GPT-4o (see Appendix~\ref{app:H5}), which considers three dimensions: parameter clarity, input constraints, and functional behavior. Each rating was given on a 1--5 ordinal scale.
\label{app:F2}

\paragraph{Score Binning.}
To reduce noise and better reflect practical groupings, we converted the 1--5 ratings into three discrete quality buckets: Low Quality (1-2), Medium Quality (3) and High Quality (4-5).

\paragraph{Inter-Annotator Agreement.}
We computed pairwise Cohen’s $\kappa$ between all annotators on the bucketed scores. Agreement scores were:
\begin{itemize}
    \item Rater 1 vs Rater 2: $\kappa = 0.536$
    \item Rater 1 vs Rater 3: $\kappa = 0.644$
    \item Rater 2 vs Rater 3: $\kappa = 0.635$
\end{itemize}
These indicate moderate to substantial agreement under standard interpretation of $\kappa$ values.

\paragraph{Correlation with GPT-4o.}
To evaluate alignment with GPT-4o, we computed Kendall’s $\tau$ between the GPT-4o scores (bucketed) and the average of human bucketed scores. The result was:
\[
\tau = 0.664, \quad p = 0.0001
\]
This demonstrates strong ordinal correlation, suggesting that GPT-4o scores are a reliable proxy for human judgments of the tool documentation quality.

\paragraph{Conclusion.}
These results support the use of GPT-4o as an automated scorer for large-scale documentation quality assessment, with acceptable agreement with human judgment and strong correlation trends.

\onecolumn
\section{Algorithm Pseudo-code}
\label{app:G}

\begin{algorithm}[H]
\caption{ToolScope: Automatic Toolset Merging and Retrieval Workflow}
\label{alg:toolscope}
\textbf{Input}: Dataset $\beta$ with queries, Initial toolset $T = \{t_1, \dots, t_n\}$ \\
\textbf{Parameter}: $k$ for Top-$k$ tools retrieved for tool selection, hybrid retrieval weight $\alpha$\\\textbf{Output}: Merged toolset $T'$, updated ground truth dataset $\beta'$, top-$k$ most relevant tools retrieved per query $q$: $T_q^{(k)}$ \\
\begin{algorithmic}[1]

\STATE \textbf{ToolScopeMerger:}
\STATE \textbf{Tool Indexing \& Filtering:} Encode tool descriptions $d_i$ into embeddings $v_i = f(d_i)$
\STATE Retrieve top-$k$ similar tools for each $t_i$ to form candidate pairs $T_i^{(k)}$
\STATE Perform cosine similarity threshold $\theta$ filtering to filter candidate pairs in $T_i^{(k)}$

\STATE \textbf{Relationship Classification:} Use LLM-based classifier $M_C$ to classify semantic equivalence for each candidate pair $(t_i, t_j)$

\STATE \textbf{Graph Construction:} Build undirected graph $G = (T, E)$ where edges represent semantically equivalent pairs
\STATE Extract connected components $C = \{C_1, \dots, C_a\}$

\STATE \textbf{Tool Pruning:} Per cluster $C_b$, create mapping of one representative tool $t_b^*$ to remaining tools in cluster $C_b$ 
\STATE Output tool mapping between tools to keep and to merge $\phi : T \to T'$

\STATE \textbf{Auto-Correction:} Use LLM validator $V$ to verify and refine each cluster $C_b \Rightarrow \{C'_1, \dots, C'_m\}$

\STATE \textbf{Toolset and Dataset Update:} For each final cluster $C'_j$, synthesize merged tool signature and description $d^*$ and update toolset: $\phi(t) \Rightarrow t_j^*, \; \forall t \in C'_j$
\STATE \textbf{return} merged toolset $T'$

\STATE Apply relabeling to benchmark dataset $\beta$ using mapping $\phi$ to produce updated dataset $\beta'$
\STATE \textbf{return} updated ground truth dataset $\beta'$
\STATE \textbf{ToolScopeRetriever:}
\FOR{each query $q \in \beta'$}
    \STATE Read decomposed queries from original query $q$ into ordered steps $S_q = \{s_1, \ldots, s_m\}$
    \FOR{each step $s \in S_q$}
        \STATE Retrieve top-$k$ candidate tools $T_s^{(k)}$ via hybrid retrieval with weight $\alpha$
        \STATE Candidates are reranked using a cross-encoder 
        \STATE Select top-$1$ tool $t^{(1)}$ based on the reranker scores 
    \ENDFOR
    \STATE Apply min-max normalization on reranker scores for all steps across their remaining tools $t^{(j)}$ for $j = 2, \dots, k$
    \STATE Select top-$(k-s)$ tools
    \STATE \textbf{return} top-$k$ tools for $q$; $T_q^{(k)}$
\ENDFOR

\end{algorithmic}
\end{algorithm}


\onecolumn
\section{Prompts}
\label{app:H}
\subsection[Prompt for Merge Auto-Correction LLM V]{Prompt for Merge Auto-Correction LLM $V$}
\label{app:H1}

\begin{figure*}[!htbp]
\small
\centering
\begin{tcolorbox}[colback=white,colframe=black,width=\textwidth,
                  boxrule=0.5pt,arc=0pt,outer arc=0pt,
                  left=6pt,right=6pt,top=6pt,bottom=6pt]

You are an expert in software tool design and resolving function overlap issues.

\textbf{Goal:} Determine whether a single candidate function is semantically similar to a given target function — not necessarily equivalent — and should be considered for potential merging or consolidation if a high degree of similarity is detected.

\textbf{Background:} Overlapping function definitions can confuse both developers and LLM-based systems when multiple similar tools compete for the same user intent. Identifying function pairs that are highly similar in behavior, purpose, and parameter usage can improve clarity, reduce redundancy, and simplify tool invocation.

\textbf{Definition of Similarity:}
Two functions are considered similar if:
\begin{itemize}
    \item They perform the same high-level task or fulfill the same user intent, even if their internal implementations differ.
    \item Their inputs and outputs are comparable or can be aligned with minimal transformation (e.g., parameters are conceptually equivalent or combinable).
    \item They could be merged into a unified function with optional parameters or internal branching, without loss of functionality.
\end{itemize}

\textbf{Examples of similarity:}
\begin{itemize}
    \item \texttt{translateToFrench(text)} and \texttt{translateToGerman(text)} $\rightarrow$ same operation (translation), different fixed parameter $\Rightarrow$ merge into \texttt{translate(text, language)}
    \item \texttt{getUserDetails()} and \texttt{fetchUserInfo()} $\rightarrow$ same intent, different naming
    \item \texttt{addCrop(crop)} and \texttt{addCropToFarm(crop, farmId)} $\rightarrow$ same core action, one more specific $\Rightarrow$ merge with optional parameter
\end{itemize}

\textbf{Not similar if:}
\begin{itemize}
    \item They serve different user intents (\texttt{validateInput} vs \texttt{sanitizeInput})
    \item They act on incompatible types or domains (\texttt{generatePDF()} vs \texttt{sendEmail()})
    \item They overlap in name but not function (\texttt{searchUser()} vs \texttt{deleteUser()})
\end{itemize}

\textbf{Input}
\begin{itemize}
    \item \textbf{Target function:} \texttt{\{target\_tool\_docstring\}}
    \item \textbf{Candidate function:} \texttt{\{candidate\_tool\_docstring\}}
\end{itemize}

\textbf{Instructions:}
\begin{itemize}
    \item Compare the target and candidate functions carefully.
    \item If they are similar based on the criteria above, return:
    \begin{itemize}
        \item The \textbf{name} of the candidate function (recommended for potential merge)
        \item A short explanation using chain-of-thought reasoning
    \end{itemize}
    \item If they are \textbf{not similar}, return:
    \begin{itemize}
        \item \texttt{None}
        \item A brief explanation of why they are not similar enough to merge
    \end{itemize}
\end{itemize}

\textbf{Output Format:}
\begin{itemize}
    \item \textbf{Line 1:} Candidate function name or \texttt{None}. Return only the name — no parameters or signature.
    \item \textbf{Line 2:} Explanation of your reasoning
\end{itemize}

\end{tcolorbox}
\caption{Prompt for LLM merging classifier $M_C$.}
\label{fig:merging_prompt}
\end{figure*}

\clearpage  
\onecolumn

\subsection[Prompt for Merge Auto-Correction LLM V]{Prompt for Merge Auto-Correction LLM $V$}
\label{app:H2}

\begin{figure*}[!htbp]
\small
\centering
\begin{tcolorbox}[colback=white,colframe=black,width=\textwidth,
                  boxrule=0.5pt,arc=0pt,outer arc=0pt,
                  left=6pt,right=6pt,top=6pt,bottom=6pt]

You are verifying whether the following Python functions can be merged into ONE API.

Similarity rubric (must satisfy 1 AND 2):
\begin{itemize}
    \item 1. Same \textit{high-level} user intent or task.
    \item 2. Inputs/outputs can be aligned by adding OPTIONAL parameters or renaming arguments (no functionality lost).
\end{itemize}

Non-merge only if INTENT clearly differs.

\textbf{Output (JSON only)}
\begin{verbatim}
{
  "merge": "MERGE_OK",
  "reason": "<optional <25 words>"
}
OR
{
  "merge": "MERGE_BAD",
  "clusters": [
    ["idA", "idB"],
    ["idC"]
  ],
  "reason": "<  <25 words >"
}
\end{verbatim}

List of candidate functions (ID : signature — doc):

\{GROUP\_BLOCK\}

\end{tcolorbox}
\caption{Prompt for Merge AutoCorrection LLM $V$.}
\label{fig:description_validator_prompt}
\end{figure*}



\subsection[Prompt for LLM Tool Documentation Merger M\_D]{Prompt for LLM Tool Documentation Merger $M_D$}
\label{app:H3}

\begin{figure*}[!htbp]
\small
\centering
\begin{tcolorbox}[colback=white,colframe=black,width=\textwidth,
                  boxrule=0.5pt,arc=0pt,outer arc=0pt,
                  left=6pt,right=6pt,top=6pt,bottom=6pt]

You are an expert Python API architect:\\
\textbf{Task:}\\
Given multiple semantically similar function definitions, merge them into
\textbf{ONE} canonical function named \texttt{\{keep\_name\}}.

Similarity rubric (all must hold to merge):
\begin{enumerate}
    \item Same user intent/task.
    \item Inputs \& outputs can be aligned via optional parameters.
    \item No functionality is lost after merge.
\end{enumerate}

\textbf{Instructions:}
\begin{itemize}
    \item Keep \texttt{\{keep\_name\}}’s parameter order \& defaults first.
    \item Add \textbf{only} unique parameters from the other functions; make them optional.
    \item Combine docstrings: start with 1–2 concise sentences of purpose, then list arguments (name, type, default, description).
    \item Do \textbf{NOT} output implementation code or markdown fences.
    \item Output \textbf{only} the final signature and full docstring.
\end{itemize}

\texttt{\{keep\_block\} \{prune\_block\}}

\end{tcolorbox}
\caption{Prompt for LLM Tool Documentation Merger $M_D$.}
\label{fig:description_merger_prompt}
\end{figure*}

\clearpage

\subsection{Prompt for LLM Agent Tool Selection}
\label{app:H4}
\begin{figure*}[!htbp]
\centering
\begin{tcolorbox}[colback=white, colframe=black, width=\textwidth,
                  boxrule=0.5pt, arc=0pt, outer arc=0pt,
                  left=6pt, right=6pt, top=6pt, bottom=6pt]

You are a tool selection agent.

Your task is to determine the most appropriate tool function (if any) to use for a specific step in a multi-step plan derived from a user request.

You are given:
\begin{itemize}
    \item The original user query to provide context
    \item A single step from the plan (typically a short sentence)
    \item A list of available tools, including their names, descriptions, argument schemas, and expected results
\end{itemize}

Your job is to return the name of the \textbf{single most appropriate function}(from the list below) that can execute this step.

Instructions:
\begin{itemize}
    \item Only output the function name exactly as written in the list below, or `None` if no tool is applicable.
    \item Do not add parentheses, arguments, or explanations.
    \item Do not make assumptions beyond what is provided in the tool descriptions and step text.
\end{itemize}

Original User Query:\\
\texttt{\{question\}}

Plan Step:\\
\texttt{\{input\}}

Candidate Tools:\\
\texttt{\{tools\}}

Output:

\end{tcolorbox}
\caption{Prompt for LLM Agent Tool Selection.}
\label{fig:description_robustness_prompt}
\end{figure*}

\subsection{Prompt for LLM Tool Documentation Quality}
\label{app:H5}
Figure~\ref{fig:description_robustness_prompt} shows the prompt for Tool Documentation Quality using an LLM.
\begin{figure*}[!htbp]
\centering
\begin{tcolorbox}[colback=white, colframe=black, width=\textwidth,
                  boxrule=0.5pt, arc=0pt, outer arc=0pt,
                  left=6pt, right=6pt, top=6pt, bottom=6pt]

\small
\textbf{You are an expert evaluator assessing the quality of tool documentation.} \\
Given a tool’s name, function signature, and docstring, decide how clear and helpful the documentation is for someone who needs to call the tool correctly.

Judge the documentation on three evidence pillars:

\begin{itemize}
  \item[\textbf{P}] \textbf{Parameters} 
    \begin{itemize}
      \item Do parameter names and counts match the signature?
      \item Are required parameters described accurately?
      \item Are optional parameters or default values acknowledged?
    \end{itemize}

  \item[\textbf{I}] \textbf{Inputs \& Constraints}
    \begin{itemize}
      \item Are input types, roles, and typical values clear?
      \item Are important constraints, ranges, or usage conditions provided?
    \end{itemize}

  \item[\textbf{B}] \textbf{Behavior \& Purpose}
    \begin{itemize}
      \item Is it clear—at least at a high level—what the tool does when invoked?
      \item Does the description indicate what the caller can expect in return (or is it self-evident)?
    \end{itemize}
\end{itemize}

\vspace{0.5em}
\hrule
\vspace{0.5em}

\textbf{Scoring rubric (1–5):}

\noindent \textbf{Score 5 – Excellent} \\
\textbf{Parameters (P):} All required and optional parameters are fully documented; naming matches the signature; defaults are stated when non-trivial. \\
\textbf{Inputs \& Constraints (I):} Types plus examples or constraints are provided for almost every field. \\
\textbf{Behavior \& Purpose (B):} Clear, precise description of functionality and return/output. \\
\textbf{Overall Quality:} Ready for production use. \\[1ex]

\noindent \textbf{Score 4 – Good} \\
\textbf{Parameters (P):} Required parameters are fully documented; most optional parameters are mentioned, with only minor detail gaps. \\
\textbf{Inputs \& Constraints (I):} Types are present, but a few examples or constraints are missing. \\
\textbf{Behavior \& Purpose (B):} Behavior is described clearly, but return/output is only briefly mentioned or assumed obvious. \\
\textbf{Overall Quality:} Very usable; small refinements desirable. \\[1ex]

\noindent \textbf{Score 3 – Fair} \\
\textbf{Parameters (P):} Required parameters are documented and correct, but several optional parameters are missing or vague. \\
\textbf{Inputs \& Constraints (I):} Basic types or roles are stated, but little constraint information is given. \\
\textbf{Behavior \& Purpose (B):} Behavior is conveyed in one or two somewhat vague sentences but is still understandable. \\
\textbf{Overall Quality:} Usable with modest guesswork. \\[1ex]

\noindent \textbf{Score 2 – Poor} \\
\textbf{Parameters (P):} Some required parameters are undocumented or mismatched with the signature. \\
\textbf{Inputs \& Constraints (I):} Types are unclear or absent; no constraint details are given. \\
\textbf{Behavior \& Purpose (B):} Behavior is barely described, and purpose is hard to infer. \\
\textbf{Overall Quality:} High risk of misuse. \\[1ex]

\noindent \textbf{Score 1 – Very Poor} \\
\textbf{Parameters (P):} Few or no parameters are documented, or documentation is misleading. \\
\textbf{Inputs \& Constraints (I):} Input expectations are absent or incorrect. \\
\textbf{Behavior \& Purpose (B):} The tool’s purpose is not conveyed. \\
\textbf{Overall Quality:} Documentation offers virtually no guidance. \\

\vspace{1em}
\textbf{Respond with exactly two lines:} \\
1. \textbf{Score} (integer 1–5) \\
2. \textbf{One-sentence justification}

\vspace{0.5em}
Now, evaluate the following tool:

\texttt{Tool name: \{tool\_name\}} \\
\texttt{Tool signature: \{tool\_signature\}} \\
\texttt{Tool description: \{tool\_description\}}

\end{tcolorbox}
\caption{Prompt for Tool Documentation Quality.}
\end{figure*}

\newpage
\section{Hyperparameters and Infrastructure}
\label{app:hyperparams}
To support reproducibility, we report here all relevant hyperparameters, thresholds, and hardware configurations used in our experiments.

\begin{table*}[!htbp]
  \centering
  \begin{tabular}{lll}
    \toprule
    \textbf{Module} & \textbf{Hyperparameter} & \textbf{Value / Notes} \\
    \midrule
    Models & Tool selection & GPT-4o, Cohere Command-R, LLaMA-3.3-70B \\
           & Embedding Model &  thenlper/gte-large \\
           & Reranker model & cross-encoder/ms-marco-MiniLM-L6-v2 \\
           & Auto-Correction model & GPT-4o \\
    \midrule
    Retrieval / Reranking & Hybrid weight \(\alpha\) & 1.0 (dense-only) \\
              Parameters & Top $m$ for reranking & 50 \\
                         & Score normalization & min–max \\
    \midrule
    Clustering & Pair generation size & 30 (top cosine neighbors) \\
               & Auto-Correction passes & 1 (single GPT-4o prompt) \\
    \midrule
    Prompts & LLM Merging Classifier & Refer to Appendix H for prompt details \\
        & Merge Auto-Correction & \\
        & Tool Documentation Merger & \\
        & LLM Agent Tool Selection & \\
        & LLM Tool Doc. Quality & \\
    \midrule
    Inference & Temperature & 0.0 \\
        & Max. Tokens & 4000 \\
        & Top $k$ & 50 \\
        & Top $p$ & 1 \\
        & Freq. Penalty & 0.0 \\
        
    \midrule
    Sensitivity & Merging threshold range & 0.77 to 0.86 \\
                         & Selected merging threshold & 0.82 \\
    \midrule
    Hardware & GPUs & 3 × NVIDIA A100 \\
    \bottomrule
  \end{tabular}
  \caption{Hyperparameter settings by module}
  \label{tab:hyperparams_modules}
\end{table*}

\subsection{Computational Costs}

The cost of running the one-time merging per tool is 1.17 cents using GPT-4o, with a latency of 3.27s per tool. On a standard A100 40 GB GPU with a single thread, for BFCL (400 tools), it completes in 22 minutes (\$4.60), for UltraTool (1.9K tools) in 2 hours (\$26), and for Seal-Tools (4K tools) in 3 hours (\$41). 

ToolScopeRetriever uses open-source embeddings and reranker with no API inference cost and latency under 1s per query. For the Tool Selection Agent, GPT-4o has 1.8s latency per row and costs 1 cent per query, while LLaMA 3.3 70B and Cohere-R-08-2024 (both open-source) have 1.5s latency and no inference cost.

\newpage
\section{Auto-Correction Case Studies (BFCL)}
\label{app:J}
To illustrate how Auto-Correction operates in practice, we provide two case studies: 
one example where semantically equivalent functions are correctly merged (\texttt{MERGE\_OK}), 
and another where an invalid merge is rejected and the cluster is split (\texttt{MERGE\_BAD}).

\subsubsection*{Case A: MERGE\_OK — Triangle area functions}
\textbf{Candidate set (ID : signature --- doc).}
\begin{quote}\small
10 : \texttt{calculate\_area(base: int, height: int, unit: str = None)} --- Calculate the area of a right-angled triangle given the lengths of its base and height.\\
0 : \texttt{calculate\_triangle\_area(base: int, height: int, unit: str = None)} --- Calculate the area of a triangle given its base and height.\\
104 : \texttt{geometry\_area\_triangle(base: int, height: int, unit: str = None)} --- Calculate the area of a triangle.\\
11 : \texttt{calculate\_triangle\_area(base: int, height: int)} --- Calculate the area of a triangle using its base and height.\\
95 : \texttt{calc\_area\_triangle(base: int, height: int)} --- Calculate the area of a triangle with the formula area = 0.
\end{quote}

\textbf{Auto-Correction output.}
\begin{quote}\small
\texttt{\{}\\
\quad \texttt{"merge": "MERGE\_OK",}\\
\quad \texttt{"reason": "All functions calculate the area of a triangle."}\\
\texttt{\}}
\end{quote}

\textbf{Rationale.}
All functions implement the same intent (\emph{triangle area from base and height}). 
Minor signature differences (e.g., optional \texttt{unit}) are reconciled. 
Docstring noise (e.g., “area = 0”) does not affect intent.

\textbf{Canonical signature (post-merge).}
\begin{quote}\small
\texttt{calculate\_triangle\_area(base: int, height: int, unit: Optional[str] = None)}
\end{quote}

\subsubsection*{Case B: MERGE\_BAD — Final velocity vs. free-fall speed}
\textbf{Candidate set (ID : signature --- doc).}
\begin{quote}\small
27 : \texttt{final\_velocity(initial\_velocity: int, acceleration: int, time: int)} --- Final velocity given initial velocity, acceleration, and time.\\
29 : \texttt{calculate\_final\_speed(time: int, initial\_speed: int = None, gravity: float = None)} --- Final speed in free fall after a certain time (neglecting air resistance).\\
31 : \texttt{calculate\_final\_velocity(initial\_velocity: int, acceleration: float, time: int)} --- Final velocity under constant acceleration.
\end{quote}

\textbf{Auto-Correction output.}
\begin{quote}\small
\texttt{\{}\\
\quad \texttt{"merge": "MERGE\_BAD",}\\
\quad \texttt{"clusters": [ ["27","31"], ["29"] ],}\\
\quad \texttt{"reason": "Function 29 assumes free fall, differing from general acceleration in 27 and 31."}\\
\texttt{\}}
\end{quote}

\textbf{Rationale.}
IDs \texttt{27} and \texttt{31} describe the same general constant-acceleration formula and can be merged. 
ID \texttt{29} assumes a free-fall case (gravity parameter), so Auto-Correction correctly split it into a separate cluster.

\newpage
\section{Context Length Reduction with ToolScopeRetriever}
\begin{table}[!htbp]
\centering
\caption{Average prompt tokens required to present the retrieved tools before and after applying ToolScopeRetriever.  
Prompt tokens are measured at $k{=}5$ for BFCL and Seal-Tools, and $k{=}30$ for UltraTool.}
\label{tab:context_reduction}

\begin{tabularx}{\linewidth}{@{}l*{3}{>{\raggedleft\arraybackslash}X}@{}}
\toprule
\textbf{Dataset} & \textbf{Original} & \textbf{Merged} & \textbf{\% Reduction} \\ 
\midrule
BFCL       & 32,563  & 469 & 98.56\% \\
Seal-Tools & 292,107 & 317 & 99.89\% \\
UltraTool  & 136,352 & 2,076 & 98.48\% \\
\bottomrule
\end{tabularx}
\end{table}
\label{app:context_length}

\section{ToolScopeMerger Merging Quality Analysis}
\subsection{Functionality Preservation of ToolScopeMerger}
\label{app:functionality_preservation}
To evaluate functionality preservation, we use two complementary metrics. \textbf{Tool-Call Coverage Rate (TCCR)} measures the fraction of gold tool calls that remain executable after merging:
\[
\text{TCCR} = \frac{1}{N} \sum_{i=1}^{N} \mathbf{1}\left[\exists\, t' \in \mathcal{T}_m \;:\; A_i \subseteq A_{t'}\right],
\]
where $N$ is the number of gold tool calls, $A_i$ is the argument set of the $i$-th call, and $\mathcal{T}_m$ denotes the merged toolset with argument schemas $A_{t'}$. TCCR is frequency-weighted, meaning frequently occurring calls contribute proportionally more.

\textbf{Unique Capability Coverage (UCC)} measures the fraction of distinct $(\text{tool}, \text{argument-set})$ capabilities that remain supported:
\[
\text{UCC} = \frac{1}{|\mathcal{C}|} \sum_{c \in \mathcal{C}} \mathbf{1}\left[\exists\, t' \in \mathcal{T}_m \;:\; A_c \subseteq A_{t'}\right],
\]
where $\mathcal{C}$ is the set of unique capabilities and $A_c$ denotes the argument set for capability $c$. UCC treats each capability equally and is therefore sensitive to rare or infrequent functionality.

A call (or capability) is considered preserved if a merged tool supports all arguments required by the original call, i.e., the original argument set is a subset of the merged tool’s schema. This ensures that merging is counted as valid only when the original functionality remains executable.

\begin{table}[h]
\centering
\begin{tabular}{lcc}
\toprule
Dataset & TCCR & UCC \\
\midrule
BFCL & 0.82 & 0.83 \\
Seal-Tools & 0.95 & 0.96 \\
UltraTool & 0.91 & 0.80 \\
\bottomrule
\end{tabular}
\caption{Functionality preservation after using ToolScopeMerger.}
\label{tab:functionality_preservation}
\end{table}

Across datasets, ToolScopeMerger preserves most functionality, with 82--95\% tool-call coverage and 80--96\% capability coverage. To specifically assess long-tail behavior, we evaluate capabilities appearing at most three times in the dataset. Preservation remains high for these infrequent capabilities: 82\% (324/395) in BFCL, 96\% (1013/1057) in Seal-Tools, and 86\% (2498/2891) in UltraTool. These results indicate that ToolScopeMerger does not disproportionately remove rare or specialized tool functionality, while maintaining strong overall executability.

\subsection{Human Evaluation of Tool Merging}
\label{app:human_eval_merges}

To further assess the reliability of ToolScopeMerger, we conducted a human evaluation on 48 merged tool clusters from the BFCL dataset. Each cluster was reviewed to determine whether the merged tools were functionally equivalent.

Results show that 95.4\% of merge decisions were correct. In addition, we evaluate the Auto-Correction module as a binary classifier for detecting and fixing incorrect merges, achieving 95.5\% precision, 93.3\% recall, and 94.4\% F1 score.

These results confirm that the merging pipeline, including the Auto-Correction stage, maintains high reliability and effectively mitigates erroneous or overly broad merges. Table~\ref{tab:human_eval_merges} summarizes these results.

\begin{table}[h]
\centering
\begin{tabular}{l c}
\hline
Metric & Score \\
\hline
Merge correctness & 95.4\% \\
Precision & 95.5\% \\
Recall & 93.3\% \\
F1 score & 94.4\% \\
\hline
\end{tabular}
\caption{Human evaluation of ToolScopeMerger on 48 BFCL tool clusters.}
\label{tab:human_eval_merges}
\end{table}

\section{ToolScopeMerger LLM Ablation}
To address concerns regarding reliance on GPT-4o in ToolScopeMerger, we conducted additional experiments using two open-source models: LLaMA 3.3 70B and LLaMA 3.1 8B as the ToolScopeMerger model. As shown in Table~\ref{tab:llm-ablation}, LLaMA 3.3 70B performs comparably to GPT-4o at $k{=}5$ (0.935 vs.\ 0.938) and slightly outperforms it at $k{=}10$ and $k{=}15$ (0.942 vs.\ 0.935; 0.945 vs.\ 0.940). LLaMA 3.1 8B also performs closely, with only a 0.016 difference in CSR across all $k$ values. These results demonstrate that ToolScopeMerger's performance is not tied to GPT-4o.

\begin{table}[h]
\centering

\begin{tabular}{lcccc}
\toprule
\textbf{ToolScopeMerger Model} & \textbf{Tool Selection Model} & \textbf{CSR@5} & \textbf{CSR@10} & \textbf{CSR@15} \\
\midrule
LLaMA 3.3 70B & GPT-4o & 0.935 & 0.942 & 0.945 \\
LLaMA 3.1 8B  & GPT-4o & 0.922 & 0.930 & 0.928 \\
GPT-4o        & GPT-4o & 0.938 & 0.935 & 0.940 \\
\bottomrule
\end{tabular}
\caption{CSR on BFCL benchmark across different ToolScopeMerger models. The Tool Selection Agent uses GPT-4o in all settings.}
\label{tab:llm-ablation}
\end{table}

\section{End-to-End System Reliability}

We conducted an end-to-end (E2E) evaluation on BFCL using 20 queries requiring merged tools across diverse domains (finance, sports, physics, law, history, and transport). Starting from the BFCL ground truth tool-parameter pairs, we generated a Python file of real, callable functions using GPT-5. We then ran ToolScopeMerger on this tool set, with a slight prompt modification to rewrite the full functionality of merged tools. The resulting merged tool set was implemented as a separate Python file of callable functions (also via GPT-5). A single ground truth file consisting of final answers was constructed and verified by 3 annotators.

Using the \texttt{smolagents} library \cite{smolagents}, we deployed a GPT-5 agent that could invoke both the original and merged function sets. The agent was run on all 20 queries across each configuration, and its final answers were compared against the ground truth. Answer correctness was assessed by GPT-5 as an LLM judge, with all judgments independently verified by 3 annotators who reached consistent conclusions. As shown in Table~\ref{tab:e2e}, ToolScope with Auto-Correction achieves 80\% final answer accuracy, substantially outperforming the baselines BM25 (30\%) and dense retrieval (40\%).

\begin{table}[h]
\centering
\begin{tabular}{lcc}
\toprule
\textbf{Method} & \textbf{Number Correct} & \textbf{E2E Accuracy} \\
\midrule
BM25                        & 6/20  & 30\% \\
Dense Retrieval             & 8/20  & 40\% \\
ToolScope + Auto-Correction & 16/20 & 80\% \\
\bottomrule
\end{tabular}
\caption{End-to-end answer accuracy on 20 BFCL queries requiring merged tools.}
\label{tab:e2e}
\end{table}

\section{Dataset Licensing}
\subsection{Seal-Tools Dataset License}

\begin{tcolorbox}[
  colframe=black,
  colback=white,
  fontupper=\ttfamily\footnotesize,
  boxrule=0.5pt,
  breakable,
  enhanced,
  sharp corners
]
The Seal-Tools dataset and associated code are licensed under the \textbf{Apache License 2.0}. Details can be found at:  
\url{https://github.com/fairyshine/Seal-Tools?tab=Apache-2.0-1-ov-file}
\end{tcolorbox}

\subsection{BFCL Dataset License}

\begin{tcolorbox}[
  colframe=black,
  colback=white,
  fontupper=\ttfamily\footnotesize,
  boxrule=0.5pt,
  breakable,
  enhanced,
  sharp corners
]
The BFCL dataset and associated code are licensed under the \textbf{Apache License 2.0}. Details can be found at:  
\url{https://github.com/ShishirPatil/gorilla/blob/main/LICENSE}
\end{tcolorbox}

\subsection{UltraTool Dataset License}
\begin{tcolorbox}[
  colframe=black,
  colback=white,
  fontupper=\ttfamily\footnotesize,
  boxrule=0.5pt,
  breakable,
  enhanced,
  sharp corners
]

The UltraTool dataset and associated code are licensed under the \textbf{Apache License 2.0}. Details can be found at:  
\url{https://github.com/JoeYing1019/UltraTool}
\end{tcolorbox}